\documentclass[10pt,journal]{IEEEtran}

\usepackage{etoolbox}
\makeatletter
\def\ps@headings{%
\def\@oddhead{\mbox{}\scriptsize\rightmark \hfil \thepage}%
\def\@evenhead{\scriptsize\thepage \hfil \leftmark\mbox{}}%
\def\@oddfoot{}%
\def\@evenfoot{}}
\makeatother
\usepackage{xcolor}
\usepackage{amsfonts}
\usepackage{mathrsfs}
\usepackage{amsfonts}
\usepackage{amssymb}
\usepackage{graphicx}
\usepackage{epsfig}
\usepackage{epstopdf}
\usepackage{psfrag}
\usepackage{mathtools}
\usepackage{array}
\usepackage{subfigure}
\usepackage{float}
\usepackage{cite,graphicx,amsmath,amssymb,color}
\usepackage{stmaryrd}
\usepackage{multirow}
\usepackage{subfig}
\usepackage{graphicx,times,amsmath} 
\usepackage{enumerate}
\usepackage{makecell}
\usepackage[justification=centering]{caption} 
\usepackage[linesnumbered,ruled,vlined]{algorithm2e}
\usepackage{lettrine}

\IEEEoverridecommandlockouts

\begin{document}
\bibliographystyle{IEEEtran}

\title{MARS: Mixed Virtual and Real Wearable Sensors for Human Activity Recognition with Multi-Domain Deep Learning Model}

\author{Ling Pei, \IEEEmembership{Senior Member,~IEEE}, \thanks{Ling Pei, Songpengcheng Xia, Lei Chu,  Fanyi Xiao, Qi Wu, Wenxian Yu and Robert Qiu are with Shanghai Key Laboratory of Navigation and Location-based Services, School of Electronic Information and Electrical Engineering, Shanghai Jiao Tong University, Shanghai, China, 200240.
Corresponding author: Lei Chu (e-mail:leochu@sjtu.edu.cn).}
Songpengcheng Xia,
Lei Chu\IEEEauthorrefmark{1}, \IEEEmembership{Member,~IEEE},
Fanyi Xiao,
Qi Wu, \\
Wenxian Yu, \IEEEmembership{Senior Member,~IEEE},
Robert Qiu, \IEEEmembership{Fellow,~IEEE}}

\markboth{Manuscript submitted to IEEE Journal}%
{Ling Pei \MakeLowercase{\textit{et al.}}: Bare Demo of IEEEtran.cls }

\maketitle
\begin{abstract}
Together with the rapid development of the Internet of Things (IoT), human activity recognition (HAR) using wearable Inertial Measurement Units (IMUs) becomes a promising technology for many research areas. Recently, deep learning-based methods pave a new way of understanding and performing analysis of the complex data in the HAR system. However, the performance of these methods is mostly based on the quality and quantity of the collected data. In this paper, we innovatively propose to build a large database based on virtual IMUs and then address technical issues by introducing a multiple-domain deep learning framework consisting of three technical parts. In the first part, we propose to learn the single-frame human activity from the noisy IMU data with hybrid convolutional neural networks (CNNs) in the semi-supervised form. For the second part, the extracted data features are fused according to the principle of uncertainty-aware consistency, which reduces the uncertainty by weighting the importance of the features.  The transfer learning is performed in the last part based on the newly released Archive of Motion Capture as Surface Shapes (AMASS) dataset, containing abundant synthetic human poses, which enhances the variety and diversity of the training dataset and is beneficial for the process of training and feature transfer in the proposed method. The efficiency and effectiveness of the proposed method have been demonstrated in the real deep inertial poser (DIP) dataset. The experimental results show that the proposed methods can surprisingly converge within a few iterations and outperform all competing methods.
\end{abstract}


\begin{IEEEkeywords}
Human activity recognition, virtual IMU, multiple-domain deep learning, uncertainty-aware consistency, transfer learning.
\end{IEEEkeywords}

\section{Introduction}

\lettrine[lines=2]{H}{UMAN} activity recognition (HAR) is a critical technology in the Internet of Things (IoT), which aims to provide information on human physical activity and detect human actions \cite{lara2012survey, kim2009human}. With the fast advancement of the IoT, HAR has attracted much attention as it enables many applications \cite{zhang2013human, pei2013human, wen2018survey, pei2018optimal} such as healthcare, security monitoring, smart home \cite{bianchi2019iot} etc. Based on different types of sensing modes, the HAR has been approached in three different ways: 1)vision-based HAR;\cite{lara2012survey, avci2010activity} 2)radio-based HAR;\cite{wang2016csi, kianoush2016device} 3)wearable-device-based HAR\cite{chavarriaga2013opportunity, zhang2019novel, xiao2020deep}. The vision-based HAR is one of the well-known external methods that has been extensively studied by academia and industry, but the vision-based HAR has to face many challenges in terms of cost, coverage, and lighting conditions, especially privacy security \cite{lara2012survey, avci2010activity}. With the deployment of wireless communication devices, the radio-based HAR used by the variations of wireless signal intensity has also been extensively studied\cite{zou2019multiple}. Besides, the radio-based HAR is vulnerable to environmental interference; The related HAR is relatively simple and is often used for location services in practical applications\cite{wang2016csi}. On the other hand, intelligent wearables equipped with inexpensive micro sensors, i.e., inertial measurement units (IMUs) \cite{chavarriaga2013opportunity}, provide a new way for HAR in a non-intrusive manner. Moreover, IMUs can be flexibly attached to the body and provide time-stamped signals with the acceleration and orientation of the body movement \cite{avci2010activity}. Therefore, in this work, we will focus on wearable-device-based HAR.

Recent years have witnessed an increasing interest in machine learning-based HAR with wearable devices. Traditional HAR algorithms, such as random forest (RF)\cite{bhanujyothi2017comparative},  Bayesian network \cite{zerrouki2018vision}, and support vector machine (SVM) \cite{pei2012using}, have been proposed for HAR based on the manual mode for choosing characteristic points. For example,   \cite{ravi2005activity, kwapisz2011activity} manually extracted different features, such as mean, energy, frequency-domain entropy and correlation, which were then fed into the classifiers, e.g., decision tree \cite{todorovski2003combining}, K-nearest neighbor (KNN) \cite{pirttikangas2006feature}. With manually selected features, the machine learning methods have achieved good recognition accuracy. Later, the advanced feature selection methods \cite{abidine2018joint}, such as principal component analysis (PCA), linear discriminate analysis (LDA), have been proposed, which can jointly minimize classification errors and decrease the computational time. 

With the rapid development of neural networks, researchers started to apply deep neural networks with HAR. Considering deep learning based HAR with  automatic feature extraction, improved accuracy and robustness have been reported in many related works \cite{zeng2014convolutional, jiang2015human, yang2015deep, hammerla2016deep, guan2017ensembles, he2018weakly, xiao2020deep}. At the early stage of deep learning based HAR, Vepakomma et al. \cite{vepakomma2015wristocracy} used traditional methods to extract manual features, which were then fed into a fully-connected deep neural network for classifying activities in HAR. As the CNN is insensitive to scale changes and can capture local dependence of nearby signal-related information, the works of \cite{zeng2014convolutional, yang2015deep, jiang2015human, hammerla2016deep} used 1D/2D CNN frameworks to perform feature extraction and classification for HAR. It has been shown in these works that the CNN is an effective model for obtaining useful information in the spatial dimension from IMU signals. In addition to using the CNN architecture to process sensor data, some researchers started to pay attention to the temporal characteristics of sensor data. For example, Guan et al. \cite{guan2017ensembles} proposed a method for HAR, which is based on the recurrent neural network (RNN)  with long short-term memory (LSTM) and obtained satisfactory results at the expense of increased computational time. What's more, the hyperbolic networks model \cite{ordonez2016deep}, consisting of convolutions and LSTM recurrent units, has been proposed to realize multi-modal wearable activity recognition. This hybrid network (denoted by Deep ConvLSTM) jointly takes into account the advantages of two kinds of network models, obtains multi-level { latent knowledge \cite{yan2019using}} for considered signals, and thus achieves better results for HAR compared with the network with single model.

In summary, the deep learning methods provide an effective way for automatic feature extraction in HAR. Besides, the hyperbolic networks with different kinds of network modules (i.e., 1D CNN, 2D CNN, RNN) encourage a flexible way for obtaining multi-level latent representation for the investigated signals. However, the performance of deep learning methods is mostly based on the quality and quantity of the collected data. There is little works on robust and efficient HAR from noisy and label-deficient IMU data.



In this study, motivated by recently published works \cite{laine2016temporal, miyato2018virtual, sun2019meta}, we propose a novel multiple-domain deep learning framework to realize robust and real-time HAR. Our contributions are listed as follows.

\begin{enumerate}
	\item We first propose to build an extensive database based on virtual IMUs. A novel multi-domain deep learning framework, employing hybrid convolutional neural networks (CNNs) in the semi-supervised form, is then proposed to learn the single-frame human activity from noisy and imbalanced data, which are very important yet rarely exploited tasks in HAR.
	\item In the proposed method, the latent representations obtained from five single-domain feature extractors are fused together according to the principle of uncertainty-aware consistency, which reduces the uncertainty by weighting the importance of the features. This is a simple yet  effective effort for improving the performance of the proposed method.
	\item { The mechanism of transfer learning is adopted in the proposed method based on the recently released AMASS dataset, containing great synthetic human poses and virtual IMU data. The employment of AMASS can enhance the variety and diversity of the training dataset and has demonstrated beneficial effects on training process and feature transfer in the proposed method.}
	\item Based on the real deep inertial poser (DIP) dataset, numerous experiments have been conducted to evaluate the performance of the problem method. The experimental results show that the proposed methods can surprisingly converge within a few iterations and outperform all competing methods, demonstrating the novelty of employing the virtual IMU and the effectiveness of the proposed methods. 
\end{enumerate}

We will release our code for reproducible research.

The remainder of this paper is structured as follows. Section \ref{sec:2} introduces the basic knowledge of data set preparation and the proposed method. Section \ref{sec:3} presents the multi-domain neural network which consists of the feature extraction module, feature fusion module and cross domain knowledge transfer module. In Section \ref{sec:4}, we evaluate the performance of the proposed method against several competing ones over various experiments. Finally, Section \ref{sec:5} concludes the work and discusses the future work.


\section{Preliminaries}
\label{sec:2}
This section introduces the data sets and related signal processing methods used for HAR tasks and presents the relevant basic knowledge. In order to ensure the diversity and richness of the HAR dataset in real scene, we innovatively reproduce a synthetic HAR dataset and generate the corresponding virtual IMU data from a recently published motion capture dataset AMASS \cite{mahmood2019amass}, which relieves the pain of scarcity of the limit training data.

\subsection{Synthesizing IMU Data}
\label{SID}

The proposed method is based on deep learning framework, in which a sufficiently large data set is required for training.  Here, we obtain synthetic IMU dataset based on the motion capture dataset AMASS \cite{mahmood2019amass} and DIP-dataset \cite{huang2018deep}, which contain abundant human activities. Therefore, according to the characteristics of selected datasets, we introduced the basic principles of the SMPL model, and briefly explained the motion capture data generation process of virtual IMU data and related data processing for labeling.

\subsubsection{Basic knowledge of SMPL model}
\label{smpl}
The SMPL model is a parametric human body model proposed by Matthew Loper \cite{loper2015smpl}, which contains 23 joint point posture descriptions of the standard bone model and a basic human body orientation description (root orientation), a total of 72 parameters describing posture (23*3+1*3), including 10 shape parameter descriptions describing the 3D human body shape. Combining the linear blend-skinning  (LBS) function $ W(\cdot) $ and the human body average template shape $ \overline{T}$ composed of $ N $ cascaded vertex vectors at the zero pose time, the basic expression of the SMPL model $M(\overrightarrow{\beta} , \overrightarrow{\theta})$ , which maps shape and pose parameters to vertices, can be obtained \cite{loper2015smpl}:
\begin{equation}
\label{smpl1}
M(\overrightarrow{\beta} , \overrightarrow{\theta}) = W(T_p(\overrightarrow{\beta} , \overrightarrow{\theta}),J(\overrightarrow{\beta}),\overrightarrow{\theta},\mathcal{W}),
\end{equation}
where
\begin{equation}
\label{smpl2}
T_p(\overrightarrow{\beta} , \overrightarrow{\theta}) = \overline{T} + B_s(\overrightarrow{\beta}) + B_p(\overrightarrow{\theta}),
\end{equation}

$ B_s(\overrightarrow{\beta})$ and $ B_s(\overrightarrow{\theta})$ are the blend shape function and the pose-dependent blend shape function, which take the body shape parameters $\overrightarrow{\beta}$ and the pose parameters $\overrightarrow{\theta}$ as input of functions respectively; $  J(\overrightarrow{\beta}) $ is the position of the joint points, which is related to the body shape parameters; $\mathcal{W}$ is a set of blend weights.

To take full advantage of the SMPL model, one need to collect real human body meshes in different poses, which are then trained for obtaining the corresponding relationship among shape parameters, pose parameters and the mesh.

\subsubsection{Motion capture dataset preparation}
The dataset containing abundant data is one of the cornerstones of deep learning research. However, general HAR datasets \cite{chavarriaga2013opportunity} only have a small amount of data and limited motion types. For the HAR task in real scene, we use a motion capture dataset commonly used for 3D reconstruction and pose reconstruction. To get the large dataset, Mahmhood and others proposed to use the Mosh++ algorithm to uniformly convert the motion capture data in different data sets into SMPL body model \cite{mahmood2019amass}, which constitutes a motion capture data set containing more than 40 hours of exercise data, more than 300 experimenters and 11,000 exercise types. The motion capture data set contained in the original AMASS data set used in this article is shown in Tab. \ref{AMASS_struct}.
\begin{table}[]
	\caption{\sc Part of The AMASS Dataset Structure}
	\label{AMASS_struct}
	\centering
	\setlength{\tabcolsep}{7mm}{
		\begin{tabular}{lllll}
			\hline
			\hline
			\multicolumn{1}{l}{\textbf{Dataset}}  & \multicolumn{1}{l}{\textbf{Motions}} & \multicolumn{1}{l}{\textbf{Minutes}} \\
			\hline
			ACCAD                                 & 252                                  & 26.74                            \\
			CMU                                   & 2083                                 & 551.56                           \\
			Eyes Japan                            & 750                                  & 363.64                           \\
			Human Eva                             & 28                                   & 8.48                             \\
			MPI HDM05                             & 215                                  & 144.54                           \\
			SSM                                   & 30                                   & 1.87                             \\
			Transitions                           & 110                                  & 15.1                             \\
			JointLimit                            & 40                                   & 24.14                            \\
			BioMotion                             & 3130                                 & 541.82                           \\
			\hline
			\hline
		\end{tabular}
	}
\end{table}

The original AMASS dataset only provides SMPL model data representing attitude parameters, and does not contain corresponding IMU data. In order to adapt to wearable gesture recognition tasks, it is necessary to simulate the IMU data of the corresponding position from the SMPL model in the AMASS dataset. Huang et al. \cite{huang2018deep} place virtual sensors on the skin surface of the SMPL model, and use forward kinematics to calculate the direction data of the virtual IMU for each frame, which is expressed in the form of a rotation matrix. For virtual simulated acceleration data, it can be obtained by two differential operations. For time $t$, the virtual acceleration can be defined as:
\begin{equation}
a_t = \frac{P_{t-1}+P_{t+1}-2P_{t}}{dt^{2}},
\end{equation}
where $P_t$ represents the location of the virtual IMU at time $t$ and $a_{t} $ represents the virtual acceleration at time $t$. As a result, we adopt the similar strategy in \cite{huang2018deep} to generate the rotation matrix and virtual acceleration data, which are further used for training the our network model.

\subsubsection{IMU data processing}

\begin{figure}[!ht]
	\centering
	\includegraphics[width=8cm]{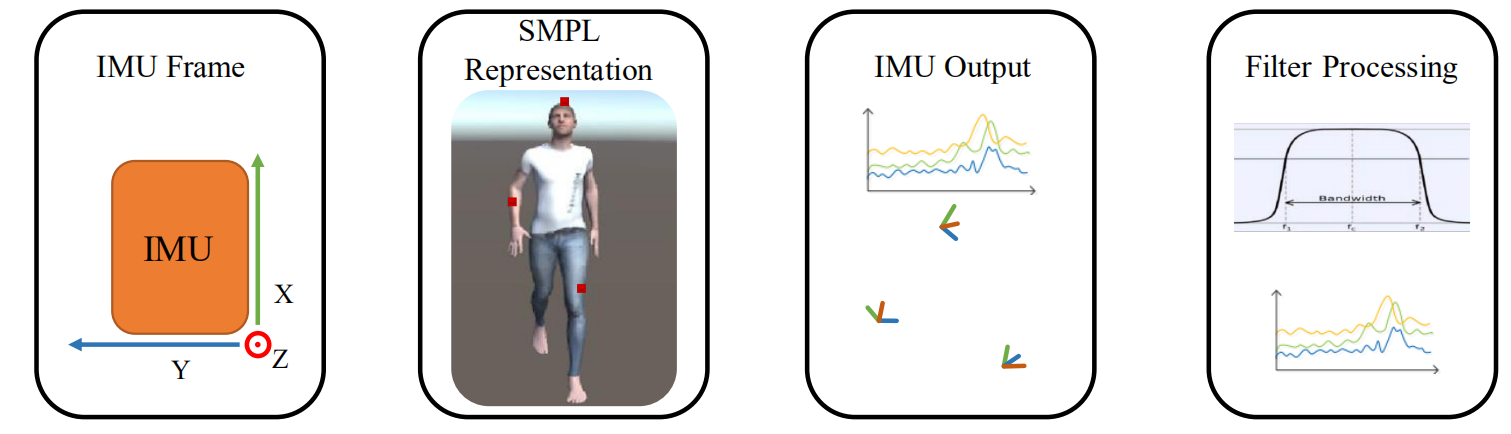}
	\caption{IMU output and SMPL body model}
	\label{IMU output and human skin model}
\end{figure}

IMU is a precision measurement device that includes an accelerometer, a gyroscope, and a magnetometer. With the development of Microelectromechanical Systems (MEMS) technology and the proliferation of IMUs in smart devices, researches on wearable IMU can better serve IoT applications. As introduced in the previous section, we use the IMU observations placed near the joints of the human body to recognize complex human activities. We use the observation values (rotation matrix and acceleration) obtained by the IMU placed near the joints of the human body to recognize the complex human activities. Since IMU sensors are susceptible to environmental noise, electromagnetic waves and temperature changes, the low-pass filter was used to reduce the impact of certain high-frequency noises in raw IMU data. Besides, in order to obtain data suitable for neural network input, we have performed necessary processing on the IMU raw data, such as coordinate conversion and normalization. Fig. \ref{IMU output and human skin model} shows a schematic diagram of the IMU output and the SMPL body model.

\subsubsection{Posture recognition data annotation} 
\begin{figure}[!ht]
	\centering
	\includegraphics[width=8cm]{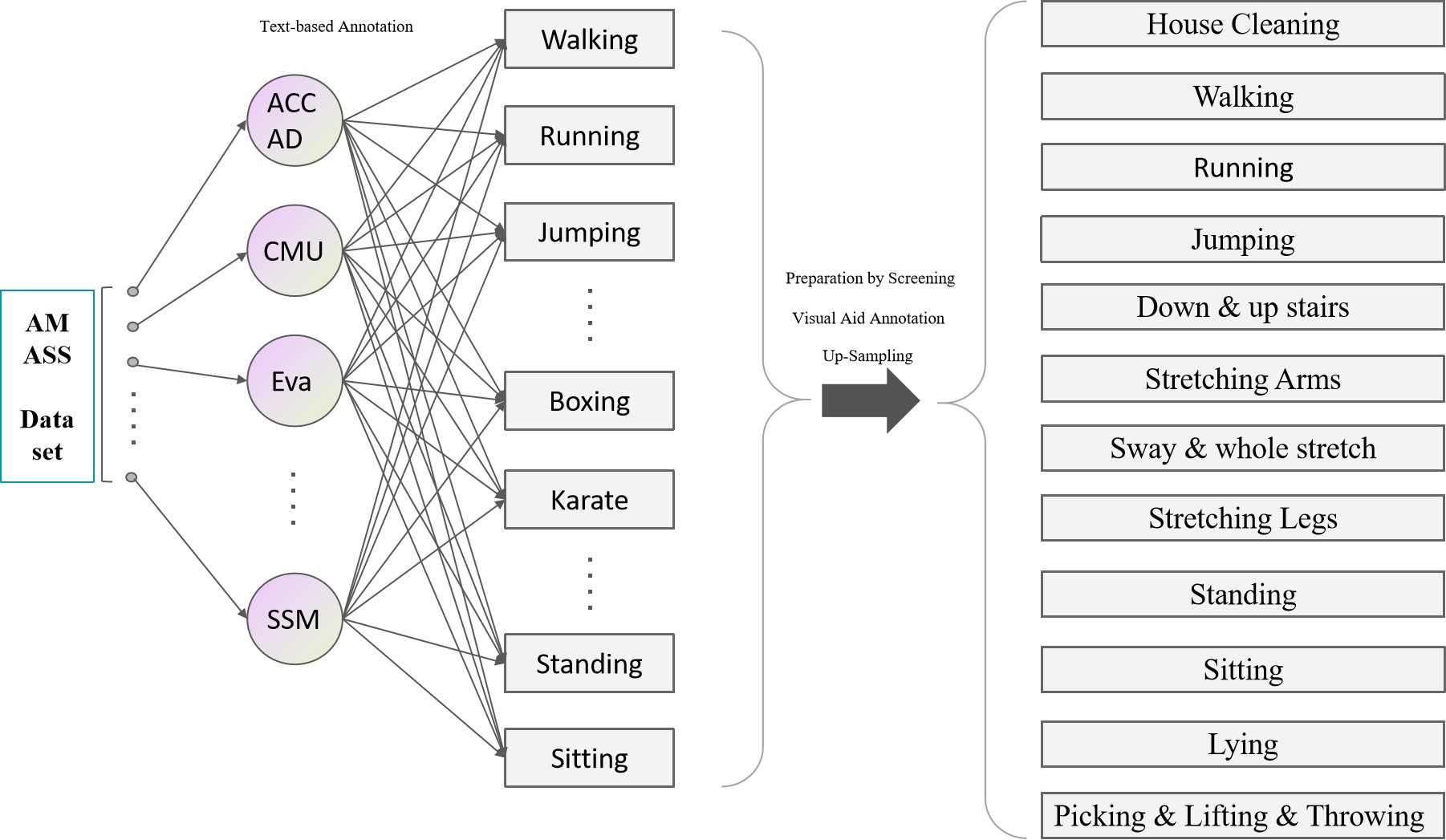}
	\caption{Data labeling flowchart.}
	\label{Data labeling flowchart}
\end{figure}

The dataset used in our work is based on the AMASS and DIP datasets as mentioned above. For the virtual IMU dataset AMASS, considering that the AMASS dataset contains a plentiful of activities, we need to screen out suitable data for HAR tasks and label them. Therefore, This paper uses the methods of text description division, common motion category screening and visual auxiliary annotation to label the HAR data of the motion capture dataset (AMASS dataset). The flowchart is shown in Fig. \ref{Data labeling flowchart}. \par

The following part introduces in detail the labeling process of the AMASS dataset: \textbf{1) \ Posture division based on text description:} The original AMASS dataset is a collection of many motion capture datasets. When collecting these data sets, the experimenter will often follow the pre-specified rules to perform the corresponding movement, so corresponding to each motion capture file, the dataset itself will provide the motion posture description corresponding to the file. This article directly uses the posture description provided by the file name as the basis to preliminarily divide the motion capture files of each sub-data set contained in AMASS.  \textbf{2) \ Common motion category screening:} After dividing the preliminary postures according to the text description, we further label them in combination with common behaviors and postures in daily life. In this process, two types of posture files, the martial arts type and specific high-frequency conversion combined postures, which are rarely related to daily human activities, were eliminated. Therefore, following the preliminary results of posture division in step 1,the AMASS dataset was finally divided into 12 HAR categories. \textbf{3) \ Auxiliary annotation of posture visualization: }After the first two steps, specific behavior classification has been completed for most data files. However, due to the lack of particular posture description information, there are still some data files that cannot be directly labeled. Therefore, we mainly use Unity and Matlab to visualize motion capture files. Therefore, after visual analysis, the data can be segmented, annotated, or eliminated, so that all the files contained in the data set are classified and annotated.
\begin{figure}[h]
	\centering
	\includegraphics[width=8cm]{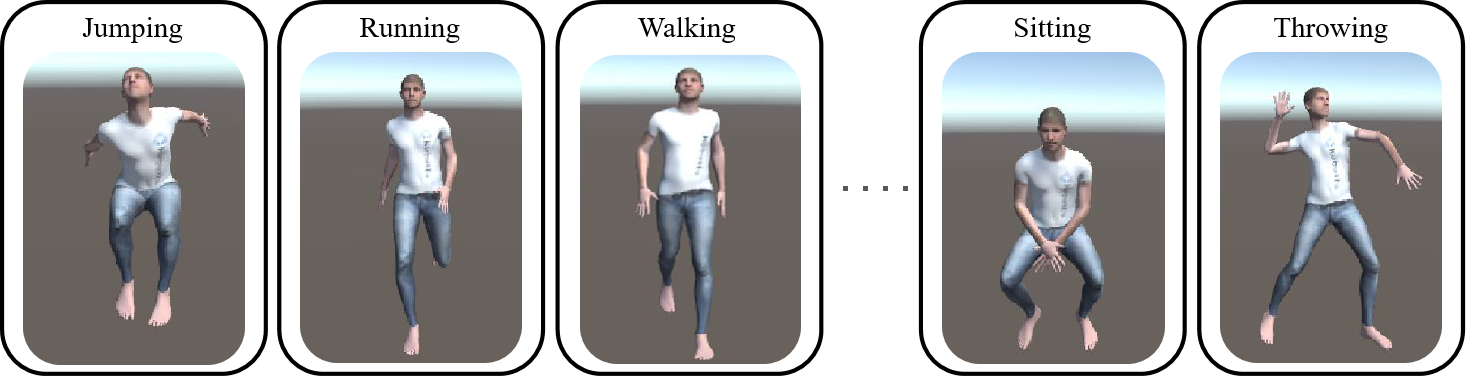}
	\caption{SMPL models with Unity visual action.}
	\label{SMPL model Unity visual action}
\end{figure}

For the DIP dataset which were collected in reality with a small amount of data, and a preliminary classification was made at the time of collection. We verified the classification of the DIP data set and eliminated some difficult-to-discern actions, and finally got 5 types of complex action categories, Upper Body, Lower Body, Locomotion, Jumping and Computer Works. Each behavior category and the specific motion posture included in each behavior category are shown in Tab. \ref{DIP_struct}.    \par
\newcommand{\tabincell}[2]{\begin{tabular}{@{}#1@{}}#2\end{tabular}}
\begin{table}[!ht]
	\caption{\sc DIP Datasets Structure}
	\label{DIP_struct}
	\centering
	\setlength{\tabcolsep}{5mm}{
		\begin{tabular}{lllll}
			\hline
			\hline
			\multicolumn{1}{l}{\textbf{Categories}}  & \multicolumn{1}{l}{\textbf{Motions}} \\
			\hline
			Upper Body  &\tabincell{c}{Arm raises, stretches and swings, Arm \\crossings on torso, and behind head.}                \\
			Lower Body  & \tabincell{c}{Leg raises. Squats. Lunges}                       \\
			Locomotion  & \tabincell{c}{Walking (Walking straight and in circle),\\ Sidesteps (crossing legs, touching feet).}       \\
			Jumping              & \tabincell{c}{Jumping jacks }                        \\
			Computer Works    & \tabincell{c}{Finishing computer works, interacting \\with everyday objects.  }                      \\
			
			\hline
			\hline
		\end{tabular}
	}
\end{table}

In what follows, we will  introduce some preliminary knowledge about denoising autoencoders and the Kullback-Leibler Divergence, which are the basis of  proposed scheme.

\subsection{Denoising Autoencoders}

With noisy observation of considered datasets (AMASS or DIP), the framework of denoising autoencoders \cite{vincent2010stacked} will be adopted to extract features and perform data reconstructions (cleaning) jointly. Given an input ${\bf{x}}$, the autoencoder-based deep neural network first encodes it to a latent space through several encoding techniques, i.e., 1D/2D convolution, pooling, and batch normalization. With similar techniques\footnote{In the decoding part, the 1D/2D convolution operations will be replaced by the deconvolution operations. In this work, following the prior work \cite{sze2017efficient} and considering the characteristics of IMU data samples, the large stride size is employed, instead of the pooling strategies, to persevere the spatial structure of the considered data.}, the features $\kappa $ embedded in the latent space are then decoded to obtain the estimation of the cleaned input ${{\bf{\hat x}}}$. The whole process can be summarized as:

\[\begin{array}{c}
{\rm{Encoding}}:\kappa  = \varphi \left( {f\left( {{{\bf{W}}_1},{\bf{x}}} \right) + {{\bf{b}}_1}} \right)\\
{\rm{Decoding}}:{\bf{\hat x}} = \varphi \left( {g\left( {{{\bf{W}}_2},\kappa } \right) + {{\bf{b}}_2}} \right)
\end{array},\]
Where $f$ and $g$ denote the techniques used in the encoding and decoding processes, respectively. In this work,  $f$ and $g$ are multiple layer convolution operations and deconvolution operation, respectively.
The goal of the autoencoder is to learn the network parameters $\left\{ {{{\bf{W}}_1},{{\bf{W}}_2},{{\bf{b}}_1},{{\bf{b}}_2}} \right\}$ by minimizing the loss function as follows,
\begin{equation}
	\label{1111}
	{\mathop {\min }\limits_{\left\{ {{{\bf{W}}_1},{{\bf{W}}_2},{{\bf{b}}_1},{{\bf{b}}_2}} \right\}} }{\sum\nolimits_{k = 1}^K {\left\| {{{{\bf{\hat x}}}_k} - {{\bf{x}}_k}} \right\|_2^2} },
\end{equation}

where $K$ denotes the number of data samples in a mini batch.


\subsubsection{Kullback-Leibler Divergence}


The Kullback-Leibler (KL) divergence, also known as the relative entropy,  is commonly used in statistics to measure similarity between two density distributions. Given two probability measures $P$ and $Q$, the KL divergence of $P$ from $Q$ can be denoted by
\begin{equation}
	\label{22222}
	{D_{KL}}\left( {P,Q} \right) = \sum {P\ln \frac{P}{Q}},
\end{equation}

In this work, we use the symmetrized version of  the to measure the difference between the learned features in the different latent spaces,
\begin{equation}
	\label{33333}
	S{D_{KL}}\left( {P,Q} \right) = {D_{KL}}\left( {P,Q} \right) + {D_{KL}}\left( {Q,P} \right),
\end{equation}
which will explained in the following section.

\section{Methods}
\label{sec:3}
\begin{figure*}[h]
	\centering
	\includegraphics[width=17cm]{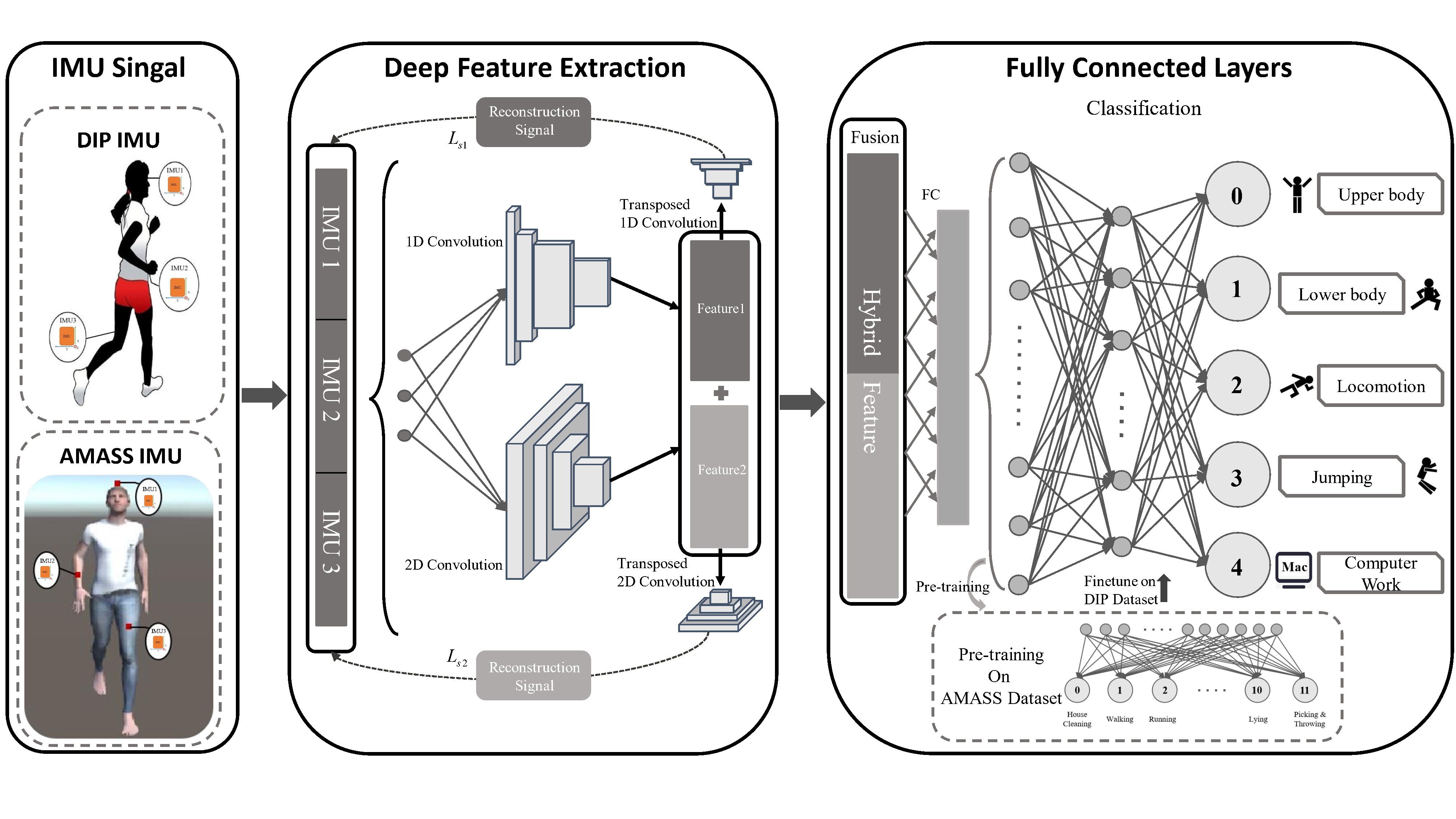}
	\caption{Proposed overview.}
	\label{Proposed Overview}
\end{figure*}

In this section, as shown in Fig. \ref{Proposed Overview}, we present the proposed multi-domain neural network which consist of three technical modules: 1) Feature extraction module based on the hybrid deep convolutional neural network; 2) Feature fusion module with the principle of uncertainty-aware consistency; 3) Cross domain knowledge transfer module according to the mechanism of the transfer learning.

\subsection{Feature Extraction using Hybrid Deep Convolutional Neural Network}
\label{femdnn}

We use the IMU data collected from three inertial sensors, which are respectively equipped on the head, left wrist and right leg of the investigated subjects. Let ${\bf{x}} \in {R^{N \times T}}$ be the collected IMU data. The symbol $N$ denotes the sum of all parameters (acceleration and the rotation matrix) and $T$ the size of the sliding window. In the following, we will elaborate on the proposed multi-domain network model for automatically extracting multi-level feature from noisy IMU signals, which are used for classifying individual human activities. We design a hybrid CNN to perform spatio-temporal feature learning of the IMU datasets.

%

As shown in Fig. \ref{Proposed Overview}, the proposed hybrid CNN model is specifically designed to identify dynamic patterns in IMU signals by incorporating feature extraction, signal denoising and classification tasks. Given the input data (DIP dataset or AMASS), ${\left\{ {{{\bf{x}}_k},{{\bf{y}}_k}} \right\}_{k = 1, \cdots ,K}}$, at the feature extraction stage, the extracted features obtained by hybrid CNNs (2D-CNNs and 1D-CNNs) can be respectively represented by
\begin{equation}
	\label{femdnn1}
	\kappa _k^{2D} = {f^{2D}}\left( {{{\bf{W}}^{2D}},{{\bf{x}}_k}} \right),
\end{equation}
and
%
\begin{equation}
\label{femdnn2}
\kappa _k^{1D} = {f^{1D}}\left( {{{\bf{W}}^{1D}},{{\bf{x}}^{T}_k}} \right),
\end{equation}
where ${f^{2D}}$ and ${f^{1D}}$ are the forward models of a multiple-layer 2D-CNN and 1D-CNN, respectively. We will present in detail about ${f^{2D}}$ and ${f^{1D}}$ in Section \ref{sec:3}. ${{\bf{W}}^{2D}}$ and ${{\bf{W}}^{1D}}$ are the corresponding parameters (weight matrices and bias vectors) in the 2D/1D networks.

The feature extraction shown in \eqref{femdnn1} and \eqref{femdnn2} jointly take advantages of 2D-CNN and 1D-CNN, which can learn the connectivity pattern among the multi-modal signals and identify characteristics in individual channel \cite{goodfellow2016deep, han2019state, karunanayake2019transfer, tang2020rethinking}, respectively. We will introduce the fusion strategies in Section \ref{EFF} for the extracted features obtained from the hybrid CNN, which take into account importance of each extracted feature.

\subsection{Extracted Features Fusion}
\label{EFF}

%

\begin{figure}[htbp]
	\centerline{\includegraphics[width=9cm]{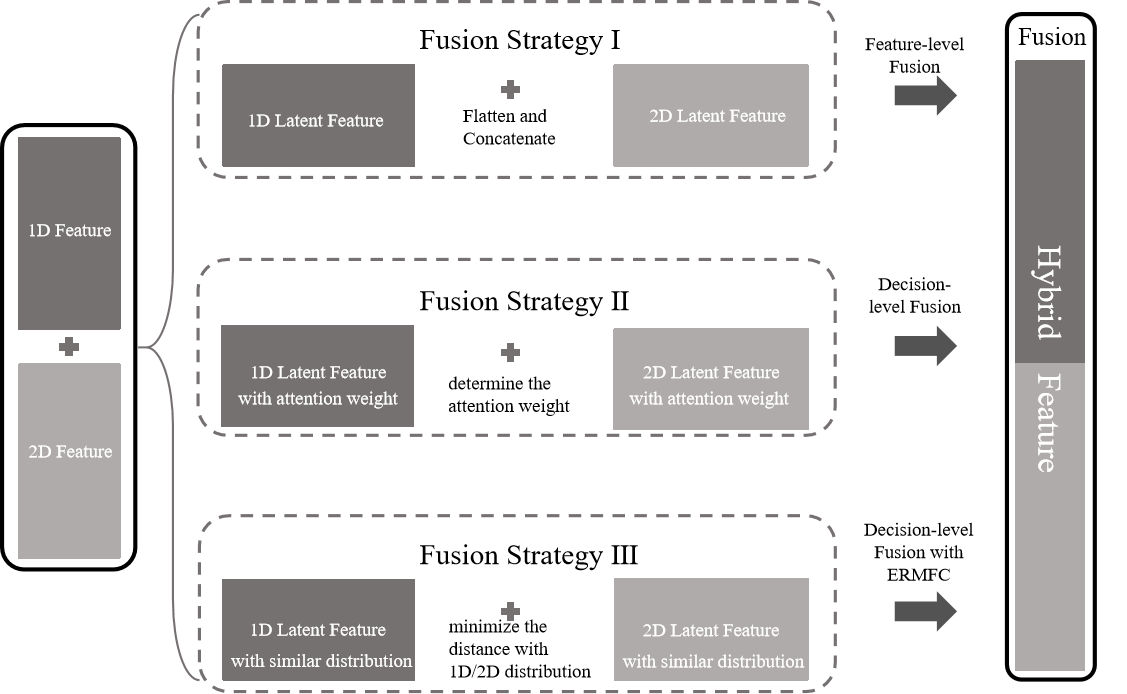}}
	\caption{Illustration of the proposed feature fusion strategies.}
	\label{fig1}
\end{figure}

In this part, as shown in Fig. \ref{fig1}, we describe the fusion strategies, such as feature-level fusion, decision-level fusion and decision-level fusion with fairness consideration, for the extracted features.

\subsubsection{Feature-level Fusion} 

As shown in \cite{yang2003feature, mangai2010survey}, it is a common way to fuse the features with the feature-level fusion scheme (or so called parallel strategy) \cite{yang2003feature}, in which the obtained features are first flattened and then concatenated into a single feature vector,
\begin{equation}
\label{EFF1}
\kappa _k^{Fussed} = {\left[ {\begin{array}{*{20}{c}}
		{\kappa _k^{2D}}&{\kappa _k^{1D}}
		\end{array}} \right]^T},
\end{equation}
which is then fed into a common classification component consisting of dense layers and the $softmax$ output layer.	The feature-level fusion scheme is a popular way for fusing the hierarchical latent representations learned by the hybrid CNN, however, ignoring the importance of the individual latent representation. In the following, we propose a decision-level fusion scheme and its improved version to address this problem.

\subsubsection{Decision-level Fusion} 

In this subsection, we introduce the design of the decision-level fusion scheme. We assume that the input (extracted feature to be fused) of the proposed fusion module is denoted as
\begin{equation}
\label{EFF2}
{\kappa _k} = {\left[ {\begin{array}{*{20}{c}}
		{\kappa _k^{2D}}&{\kappa _k^{1D}}
		\end{array}} \right]^T} \in {R^{M \times K}},
\end{equation}
where $M$ is the dimension to be merged and $K$ is the output dimension of each merging feature. The general expression of the decision-level fusion scheme can be formulated as
\[\begin{array}{*{20}{c}}
{{\xi _k} \odot {\kappa _k}}&{{\rm{s}}{\rm{.t}}{\rm{. }}\sum {{\xi _k} = 1},}
\end{array}, \ 0 \le {\xi _k} \le 1,\]
where ${{\xi _k}}$ is the attention weights that control the contributions of each merging feature in ${\kappa _k}$. As a result, the key problem now is to find the way to determine the attention weights.
Here, we propose to use the $sigmoid$ function ($\sigma \left(  \cdot  \right)$) to satisfy the condition in
\eqref{EFF2}. As a result, we can have
\begin{equation}
\label{EFF3}
\kappa _i^{Fussed} = \sqrt {\sigma \left( {\kappa _i^{2D}} \right)}  \odot \kappa _i^{2D} + \sqrt {1 - \sigma \left( {\kappa _i^{1D}} \right)}  \odot \kappa _i^{1D},
\end{equation}
where $\odot$ denotes the Hadamard product.

In order to further improve the stability of the proposed decision-level fusion scheme, we enable the design to perform the fusion with empirical risk minimization under fairness constraints (ERMFC) \cite{sugiyama2007covariate, donini2018empirical}.

\subsubsection{Decision-level Fusion with ERMFC} The fusion strategy \eqref{EFF3} does not fully consider the fairness of the extracted feature i.e., the size and the distribution of each feature. Following the theoretical works shown in  \cite{sugiyama2007covariate, donini2018empirical}, the fairness of the extracted feature will be enhanced when they are sampled from the same distribution. Such a condition is extremely hard to achieve for a network functional. As a result, we try to tackle the problem by minimizing the distance between their distribution as
\begin{equation}
\label{fusion3}
{DKL\left( {\kappa _i^{2D},\kappa _i^{1D}} \right) = {D_{KL}}\left( {\kappa _i^{1D},\kappa _i^{2D}} \right) + {D_{KL}}\left( {\kappa _i^{2D},\kappa _i^{1D}} \right)},
\end{equation}
where
\[\kappa _i^{1D} = \frac{{{\kappa ^{1D}}}}{{\sum {{\kappa ^{1D}}} }},{\kappa ^{1D}} = \frac{1}{{{N_1}}}\sum\limits_{i = 1}^{{N_1}} {\kappa _i^{1D}} ,\]
\[\kappa _i^{2D} = \frac{{{\kappa ^{2D}}}}{{\sum {{\kappa ^{2D}}} }},{\kappa ^{2D}} = \frac{1}{{{N_1}}}\sum\limits_{i = 1}^{{N_1}} {\kappa _i^{2D}}. \]

Once the network was trained and the optimal condition of network optimization was achieved, the fairness among different features will be achieved at optimum \cite{sugiyama2007covariate, donini2018empirical}.

With fine-grained features as shown above, we will use standard Softmax regression,
\begin{equation}
\label{fusion4}
{J_s} = \mathop {\min }\limits  \left( { - \frac{1}{M}\sum\limits_{i = 1}^M {\sum\limits_{j = 1}^K {{1_{\left\{ {{y_i} = j} \right\}}}\log \frac{{{e^{{{\left( {\kappa _j^{Fussed}} \right)}^T}{{\bf{x}}_i}}}}}{{\sum\limits_{j = 1}^K {{e^{{{\left( {\kappa _j^{Fussed}} \right)}^T}{{\bf{x}}_i}}}} }}} } } \right),
\end{equation}
to train the network at the training stage and finally obtain the estimates for the HAR activities. In \eqref{fusion4}, $1(.)$ denotes the indicator function whose value will be 1 if the condition ${{y_i} = j}$ is satisfied, otherwise 0.

\subsection{Signal Denoising in Two Pathways}
\label{sdtp}
To reduce the effect of noise that inevitably contained in the IMU signals, we are trying to reconstruct the IMU signals in two pathways by solving the following minimization function:
\begin{equation}
\label{sdtp1}
{J_r} = \begin{array}{*{20}{c}}
{\mathop {\min }\limits_{\bf{W}} }&{\frac{1}{2}\sum\nolimits_{k = 1}^K {\left\| {{\bf{\hat x}}_k^{1D} - {\bf{x}}_k^{T}} \right\|_2^2} }
+ \sum\nolimits_{k = 1}^K {\left\| {{\bf{\hat x}}_k^{2D} - {\bf{x}}_k} \right\|_2^2} \end{array},
\end{equation}
where
\[{\bf{\hat x}}_k^{1D} = {g^{1D}} {\left( {{{\left( {{{\bf{W}}^{1D}}} \right)}^{'}},\kappa _i^{1D}} \right)}\] and \[{\bf{\hat x}}_k^{2D} = {g^{2D}}\left(  {{{\left( {{{\bf{W}}^{2D}}} \right)}^{'}},\kappa _i^{2D}} \right)\] are outputs (reconstructed signals) of the proposed  2D-CNNs and 1D-CNNs, respectively. The functions ${g^{2D}}$ and ${g^{1D}}$ denote the models of a multiple-layer 2D-DCNN and 1D-DCNN, respectively. We will present in detail about ${g^{2D}}$ and ${g^{1D}}$ in Section \ref{sec:3}. ${\bf{W}} = \left\{ {{{\left( {{{\bf{W}}^{2D}}} \right)}^{'}},{{\left( {{{\bf{W}}^{1D}}} \right)}^{'}}} \right\}$,  are the corresponding parameters (weight matrices and bias vectors) in the 2D/1D networks.
The design in \eqref{sdtp1} enables us to denoise the IMU signals in two distinct pathways, improving the robustness of the proposed algorithm.

\subsection{Overall Network Parameters Optimization}

With above technical analysis, we obtain the final objective function which acts as the general rule for the network parameters learning as
\begin{equation}
\label{onpo_1}
\begin{array}{*{20}{c}}
{\mathop {\min }\limits_{\bf{W}} }&{{J_s} + {\beta _1}{J_r} + {\beta _2}\Psi \left( {\bf{W}} \right)}
\end{array} + {\beta _3}DKL\left( {\kappa _i^{2D},\kappa _i^{1D}} \right),
\end{equation}
where ${\bf{W}}$ are network parameters in all considered networks and ${\left\{ {{\beta _k}} \right\}_{k = 1,2,3}}$ are auxiliary variables that are used for balancing each penalty in \eqref{onpo_1}. ${\Psi \left( {\bf{W}} \right)}$ is an regularization on the model parameters (${{{\bf{W}}_i}}$ are the ensembles of $i$-th network weights and bias), which is defined as follows,
\begin{equation}
\label{23434}
\Psi \left( {\bf{W}} \right) = \sum {{{\left\| {{{\bf{W}}_i}} \right\|}^2}},
\end{equation}

The network parameters will be updated using the stochastic gradient descent strategy \cite{lecun2015deep} with a dynamic learning rate scheme, which will be elaborated in the following section.

\subsection{Knowledge Transfer from Synthetic to Real IMU Data}

\begin{algorithm}[!t]
	\DontPrintSemicolon
	\KwIn{AMASS: $\left\{ {{\bf{x}}_{train}^{A},{\bf{y}}_{train}^{A}} \right\}$,  $\left\{ {{\bf{x}}_{eval}^{A},{\bf{y}}_{eval}^{A}} \right\}$;
		\\ DIP:  $\left\{ {{\bf{x}}_{train}^{D},{\bf{y}}_{train}^{D}} \right\}$,  $\left\{ {{\bf{x}}_{eval}^{D}} \right\}$}
	\KwOut{Activity estimates for DIP: ${{\bf{\hat y}}_{eval}^A}$}
	Initialize the weights and bias (${\bf W}$) using Xavier initialization; \\
	$i \gets 0, lr \gets 0.001, factor = 0.99 $\;
	\While{$i \leq max_{iter}$ or {Convergent}}{
		Get mini batch data in AMASS;\\		
		$i \gets i + 1$\;
		Update ${\bf W}$ according to \eqref{onpo_1}; \\
		\If{mod(i, 100) = 0}{$lr = lr*factor$}
		Calculate ${{\bf{\hat y}}_{eval}^A}$ according to  \eqref{fusion4} and examine convergence;				
	}
	\Return{${{\bf{\hat W}}}$}\;
	
	Initialize the weights and bias in fusion layer using Xavier initialization; \\
	Initialize the weights and bias in other layers with ${{\bf{\hat W}}}$; \\	
	\While{$i \leq max_{iter}$ or {Convergent}}{
		Get mini batch data in DIP;\\	
		$i \gets i + 1$\;
		Update ${\bf W}$ according to \eqref{onpo_1};
	}
	Calculate ${{\bf{\hat y}}_{eval}^D}$ according to  \eqref{fusion4};
	\Return{${{\bf{\hat y}}_{eval}^D}$}\;
	\caption{{\sc Knowledge Transfer Algorithm}}
	\label{alg1}
\end{algorithm}

As shown in Section \ref{SID}, we have created a very large synthetic IMU data (AMASS) based on the recently released AMASS dataset, containing abundant human poses. The knowledge transfer from synthetic IMU data to the real one is enabled by the mechanism of the transfer learning \cite{torrey2010transfer}, which is summarized in Algorithm. \ref{alg1}.

So far, we have introduced the design of the proposed network model. We will evaluate its performance with numerous experiments in the following.  
%
%
%
%

\section{Implementations and Discussions}
\label{ID}

Combining the related theories of feature extraction, fusion and signal reconstruction introduced earlier, we now briefly explain the network parameters and structure. 

\subsection{Network parameters specification}

The network architecture consists of five single-domain modules: a 1D-CNN encoder, a 1D-CNN decoder, a 2D-CNN encoder, a 2D-CNN decoder and a feature fusion classifier.

In the first module, the input IMU data in the mini-batch has a dimension of $36\times60$, which means the data in the time-domain were assigned in the sub-channels, leading to effective extraction of latent feature in the spatial-domain. Besides, the 1D-CNN encoder has four 1-strided 1D-convolutional layers, which reduce the input size from 60 to 48, and change the channel from 36 to 128. Next, the convolutional feature of dimension $128\times48$ were flattened into $6144\times1$ to obtain 1D latent features. The 1D-CNN decoder contains four 1-strided transposed 1D-convolutional layers, which restore the IMU data from 48 to 60, and change the channel from 128 to 36. Before passing the 4 transpose layers, the feature data needs to be reshaped to $64\times128$. The 1D-CNN module and its transposed version comprise first pathway for IMU data construction.

The second signal reconstruction way includes two network single-domain modules: a four-layer 2D-CNN encoder and a four-layer 2D-CNN decoder.
The input has a dimension of $1\times60\times36$, which means the input IMU data were allocated in one sub-channel to obtain the connectivity pattern of all sensors in one time window.
The 2D-CNN encoder has two 2-strided 2D-convolutional layers and two 1-strided 2D-convolutional layers, which reduce the input size from $60\times36$ to $7\times1$, and change the channel number form 1 to 96. Next,  the convolutional feature data of size $96\times7\times1$ were reshaped into $672\times1$ to obtain flattened 2D features. The 2D-CNN decoder contains two 1-strided transposed 2D-convolutional layers and two 2-strided transposed 2D-convolutional layers, which restore the IMU data from  $7\times1$ to $60\times36$, and change the channel number from 96 to 1. These two signal reconstruction ways were connected by a feature fusion network module, which has three kinds of forms (See detailed introductions in Section \ref{EFF}). This module takes hybrid (1D/2D) latent feature as input, and is then fed into fully connected layers to obtain the HAR activity  estimates.


\subsection{Convergence Performance}
\label{CP}
\begin{figure}[h]
    \centering
    \includegraphics[width=8cm]{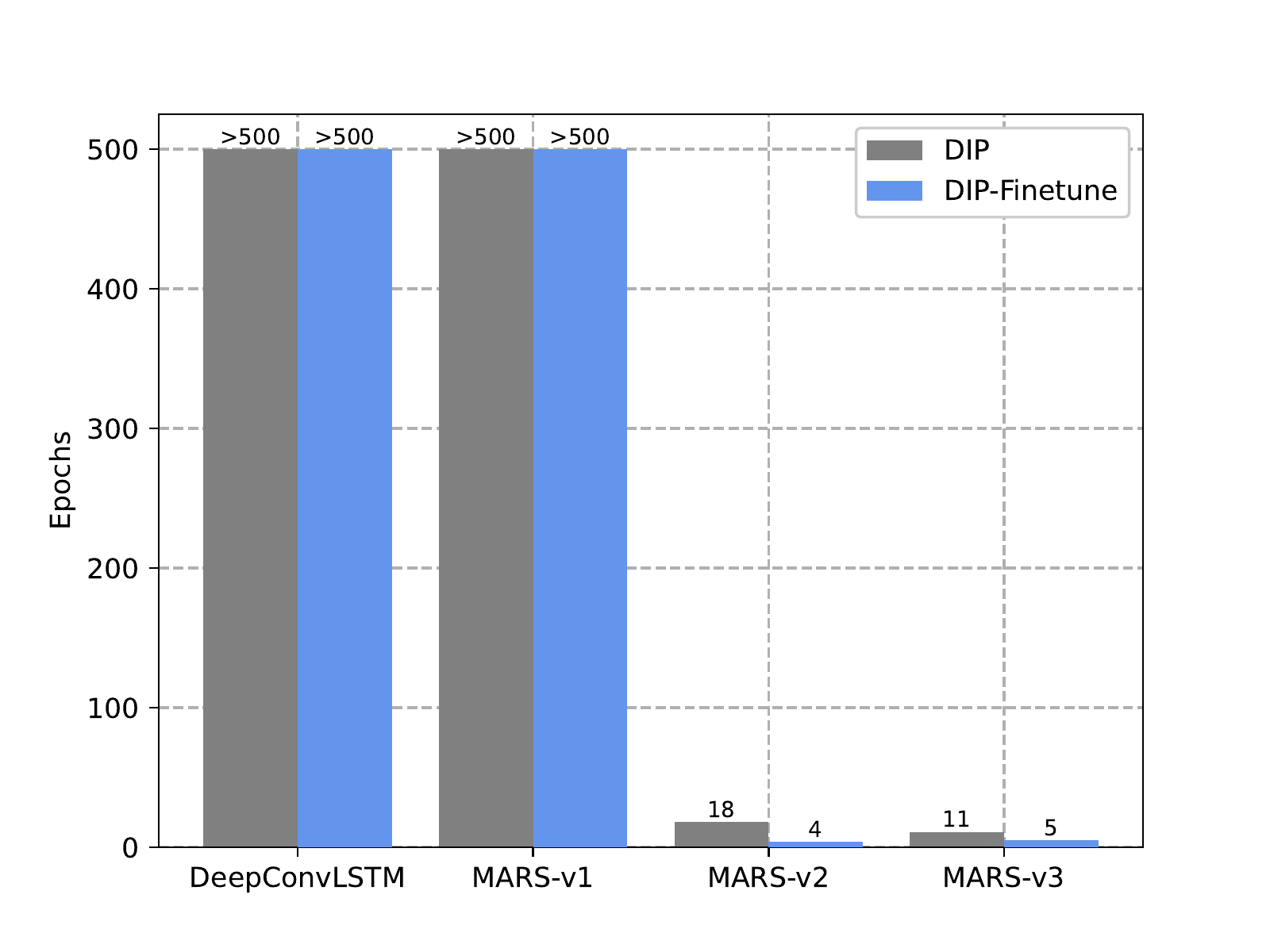}
    \caption{Convergence performance comparison.}
    \label{shoulian}
\end{figure}

This part discusses complexities of the proposed network model in terms of its training performance. To compare the convergence speed of different methods, we define convergence as the accuracy of training during the training process reaches 99\% and remains stable for 10 epochs. In Fig. \ref{shoulian}, we compare the convergence performance of the three methods proposed in this article and a representative competing algorithm (Deep ConvLSTM). The proposed methods with three different kinds of fusion strategies are denoted by MARS-v1, MARS-v2, and MARS-v3, respectively.

In the experiment, we set up to 500 epochs. Among them, the training accuracy of the method MARS-v1 and the comparison algorithm Deep ConvLSTM cannot reach the set conditions in the upper limit of 500 iterations. The MARS-v2 and MARS-v3 can converge at a surprising speed whether they are directly trained on the DIP dataset or fine-tuned by the AMASS dataset pre-training model. And using the model pre-trained on the AMASS data set, through fine-tuning training on the DIP data set, it can converge only after 4 iterations. This result also shows that the proposed method has fast convergence characteristics, and the use of the pre-trained model on the AMASS data set for fine-tuning can converge in just a few iterations, which is of great significance in practical applications.

\begin{table}[]
	\caption{\sc Train Convergence Performance Comparison}
	\label{shoulian_compare}
	\setlength{\tabcolsep}{1.2mm}{
	\begin{tabular}{lclclclclcl}
	    \hline
		\hline
		        &   \multirow{2}{*}{$Settings$}    & \multicolumn{2}{c}{\textbf{$DIP$}}     & \multicolumn{2}{c}{\textbf{$Finetune$}}          \\ 
		\cline{3-4}   \cline{5-6}
		&            & $10 Epoch $   & $200 Epoch$    & $10 Epoch$     & $200 Epoch$     \\
		\hline 
		& \textbf{Deep ConvLSTM} & 35.90\% & 94.97\%  & 93.11\%  &  92.69\% \\
		\hline
		& \textbf{MARS-v1} & 97.56\% &  97.81\%  & 97.96\% & 97.98\% \\
		\hline 
		& \textbf{MARS-v2} & 97.99\% & 99.96\%  & 99.96\%  &  100.00\% \\
		\hline
		& \textbf{MARS-v3} & 97.51\% &  100.00\%  & 99.95\% & 100.00\% \\
		\hline 
		\hline
	\end{tabular}
	}
	
\end{table}

Tab. \ref{shoulian_compare} further shows the convergence characteristics of the four methods. We compared the training accuracy of 10 epochs and 200 epochs. For Deep ConvLSTM and MARS-v1 whose training accuracy does not reach 99\% after 500 Epochs, Fig. \ref{shoulian_compare} shows that the training accuracy of all proposed methods are higher than Deep ConvLSTM under the same number of training epoch. Moreover, it is shown in Fig. \ref{shoulian_compare} that the training accuracy of all methods with knowledge transfer ( Algorithm. \ref{alg1}) can obtain satisfactory accuracy within several training epochs, which demonstrates that the efficiency of the proposed methods.

\subsection{Discussions}

{ The general objective of HAR based on handy instruments, i.e., IMUs, is to obtain knowledge about human physical activity and then achieve intelligent control, such as smart healthcare and smart home. It is noted that the multi-modality in the non-stationary signals collected by IMUs makes it a hard way to realize robust data analysis. Besides, the scarcity of informative information aggravates pain in achieving accurate HAR. On the other hand, the deep learning-based methods pave the way for better understanding the signals. We have proposed a novel multi-domain deep learning framework through the above analysis, enabling automatic and efficient feature extraction from both the synthetic and real data.

From the perspective of practical implementation, the computational complexity is a significant concern for many platforms with limited computing resources. In this work, we first trained the network with a large-size AMASS dataset, which can be completed in a control center with rich computing resources. The network model has a medium-size (about 20M) and was saved. As shown in Section \ref{CP}, with the pre-trained network parameters, experimental results on the real DIP dataset demonstrated that the proposed method could surprisingly obtain a target accuracy within few iterations, indicating the novelty of employing the virtual IMU and the effectiveness of the proposed methods. We believe this work will be beneficial for both researchers and practitioners.
}

\section{Experiment Results}
\label{sec:4}
In this section, we will introduce the experimental setup and evaluation results of the proposed method.  First, we briefly describe the data set and experimental settings used in the experiment; second, we evaluate the effect of the algorithm in public datasets, and then present the experimental results of the proposed methods and several competing HAR ones. All algorithms are performed on the Pytorch platform in a workstation equipped with an Intel i7-8750 processor, 32GB memory, and eight NVIDIA GTX 1080 Ti GPUs.

\subsection{Experimental Setup}
In order to evaluate the effectiveness of the proposed model, we conducted extensive experiments on the datasets:AMASS and DIP, which have been introduced in Section \ref{sec:2}. It is worth noting that the IMU data in AMASS is virtual IMU data, which is obviously different from real IMU data. Therefore, in order to improve the versatility of the data set in the real scene, all our methods are pre-trained on AMASS, and then using transfer learning to fine-tune the network on the DIP data set, and finally test on the DIP datasets.

\subsection{Performance Metrics}

In this paper, three performance metrics \cite{torrey2010transfer, lecun2015deep}: accuracy, precision, f1-score, are employed to evaluate the final classification performance of our proposed method in HAR.

$\textbf{Accuracy}$: Accuracy is the overall accuracy for all classes calculated as
\begin{equation}
\begin{split}
Accuracy = \frac{\sum_{i=1}^{N}TP_{i}+\sum_{i=1}^{N}TN_{i}}{\splitfrac{\sum_{i=1}^{N}TP_{i}+\sum_{i=1}^{N}TN_{i}+}{\sum_{i=1}^{N}FP_{i}+\sum_{i=1}^{N}FN_{i}}},
\end{split}
\end{equation}
where $N$ denotes the class number. Variables $TP_{i}$, $FP_{i}$, $TN_{i}$, $FN_{i}$ are the true positives, false positives, true negatives and false negatives of the class $i$, respectively.

$\textbf{Precision}$: Precision is the precision of correctly classified positive instances to the total number of instances classified as positive, which can be denoted by
\begin{equation}
Precision =\frac{1}{N} \sum_{i=1}^{N} \frac{TP_{i}}{TP_{i}+FP_{i}}.
\end{equation}

$\textbf{F1-score}$: F1-score is a key evaluation measure of classification performance, which considers both the precision and the recall of the test:
\begin{equation}F1_{score}=\frac{2 \sum_{i=1}^{N} TP_{i}}{2 \sum_{i=1}^{N} TP_{i}+\sum_{i=1}^{N} FP_{i}+\sum_{i=1}^{N} FN_{i}}.
\end{equation}

\subsection{ Compared Algorithms}

We compare the three proposed methods with the following algorithms: \par
$\textbf{Random forest (RF)}$ \cite{liaw2002classification}: it refers to a classifier that constructs a multitude of decision trees to train and predict samples. \par
$\textbf{LSTM}$  \cite{zebin2018human}: A multi-layer RNN network with long short-term memory has been proposed for HAR based on inertial sensor time-series. \par
$\textbf{Deep ConvLSTM}$  \cite{ordonez2016deep}: This model uses a deep convolutional neural network to extract feature and a recurrent neural network to learn time dependencies. \par
$\textbf{Xiao2019}:$ In \cite{xiao2020deep}, a deep convolutional autoencoder based network have been proposed for HAR based on the DIP dataset.

\subsection{Overall Performance Comparison}
\label{opc}
In this experiment, we select the data of three inertial sensors at the head, left wrist and right leg for training and testing. The specific data organization is shown in Tab. \ref{3IMUform}. Where Acc represents acceleration, Ori represents the rotation matrix measured by IMU.

\begin{table}[]
	\caption{\sc IMU Data Organization Form}
	\label{3IMUform}
	\begin{tabular}{llllll}
		\hline
		\hline
		Acc\_head                    & Ori\_head                     & Acc\_wrist                     & Ori\_wrist                     & Acc\_knee                      & Ori\_knee                      \\ \hline
		\multicolumn{1}{c}{1$\sim$3} & \multicolumn{1}{c}{4$\sim$12} & \multicolumn{1}{c}{13$\sim$15} & \multicolumn{1}{c}{16$\sim$24} & \multicolumn{1}{c}{25$\sim$27} & \multicolumn{1}{c}{28$\sim$36} \\ \hline
		\hline
	\end{tabular}
\end{table}

For the AMASS dataset and DIP dataset, we organize them according to the format as shown in Tab. \ref{3IMUform}. Besides, we use the AMASS dataset for training to obtain pre-training models of all employed networks. Then, we use the pre-trained model on the AMASS data set to initialize the new network weights, and then perform the knowledge transfer according to Algorithm. \ref{alg1} for the DIP dataset and get the test results.

%


 \begin{figure*}[!ht]
 	\subfigure[The result of MARS-v2 without fine-tuning]{
 		\includegraphics[width=8cm]{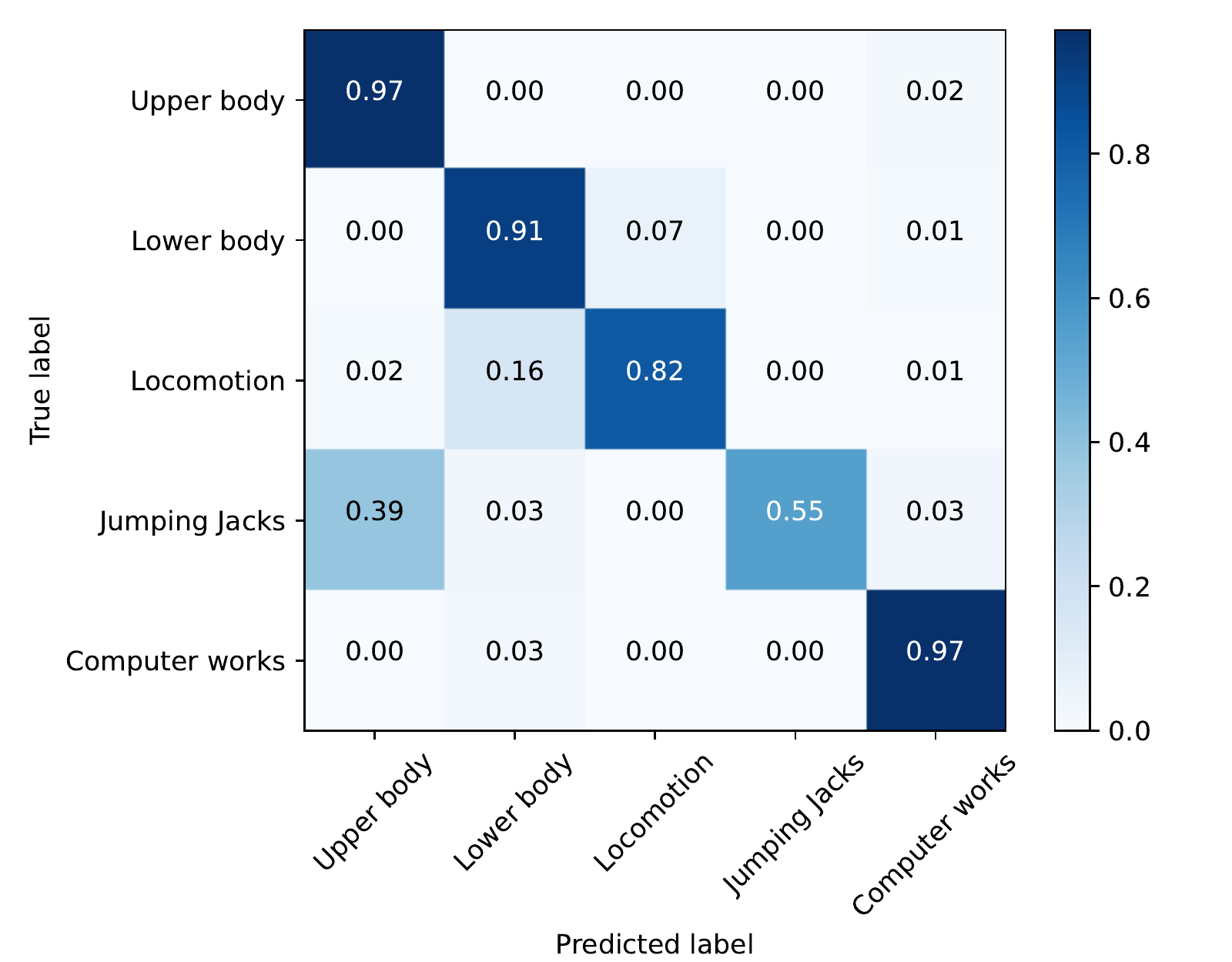}}
 	\subfigure[The result of MARS-v2 with fine-tuning]{
 		\includegraphics[width=8cm]{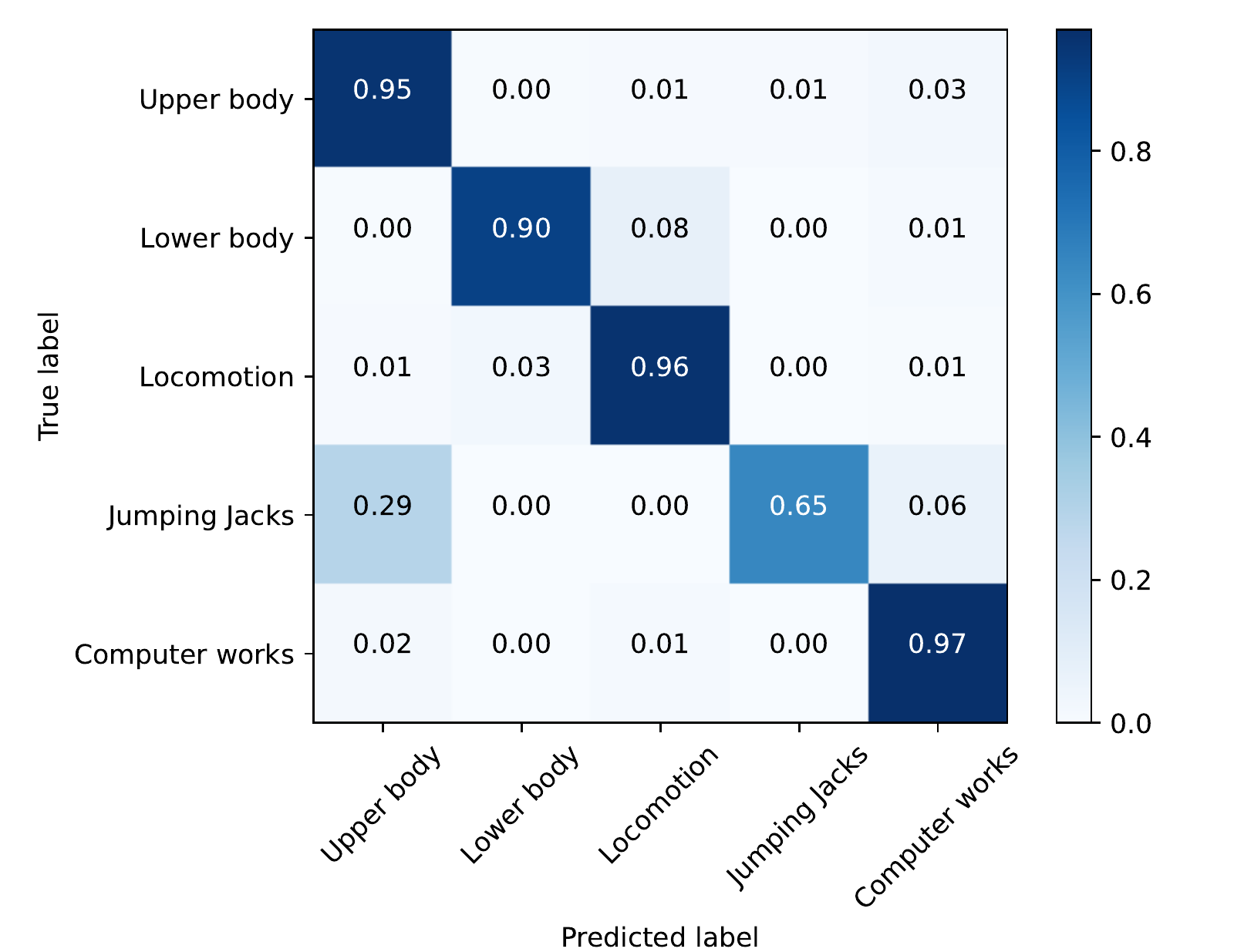}}   \\
 	
 	\subfigure[The result of MARS-v3 without fine-tuning]{
 		\includegraphics[width=8cm]{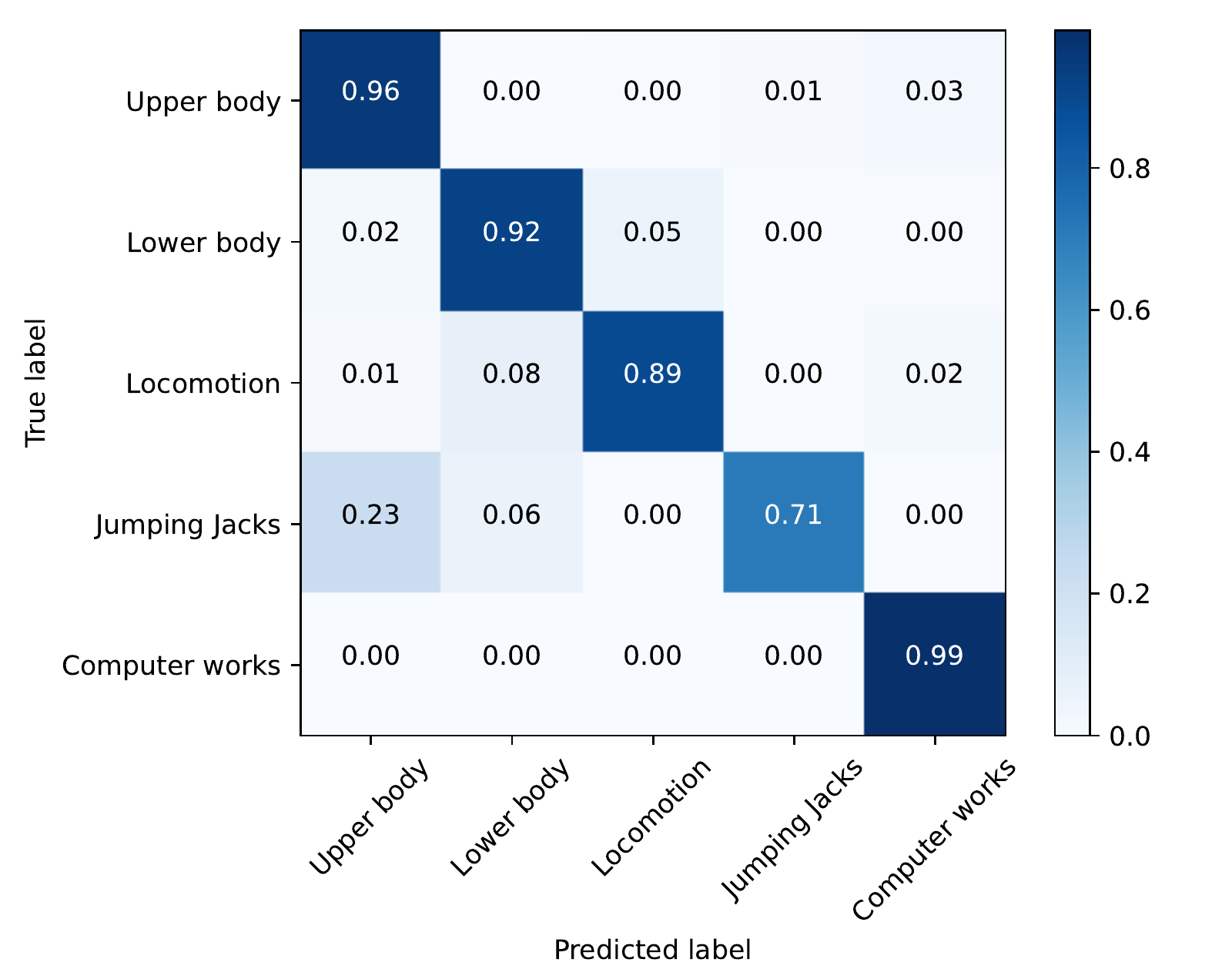}}
 	\subfigure[The result of MARS-v3 with fine-tuning]{
 		\includegraphics[width=8cm]{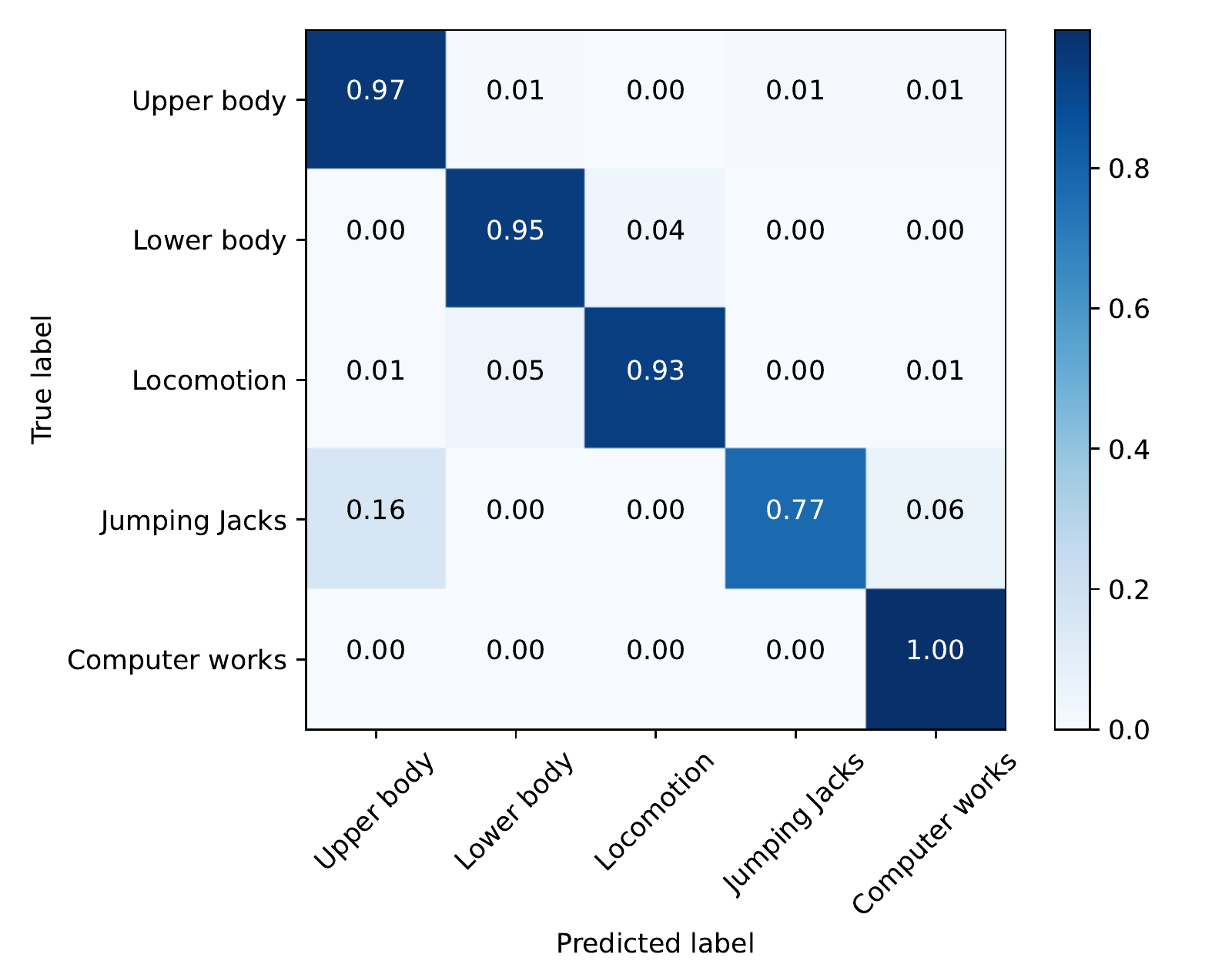}}
 	
 	\caption{Confusion matrices illustration in the first case (3 IMUs).}
 	\label{level}
 \end{figure*}
 
 \begin{figure*}[] 
	\subfigure[MARS-v1]{
		\includegraphics[width=6cm]{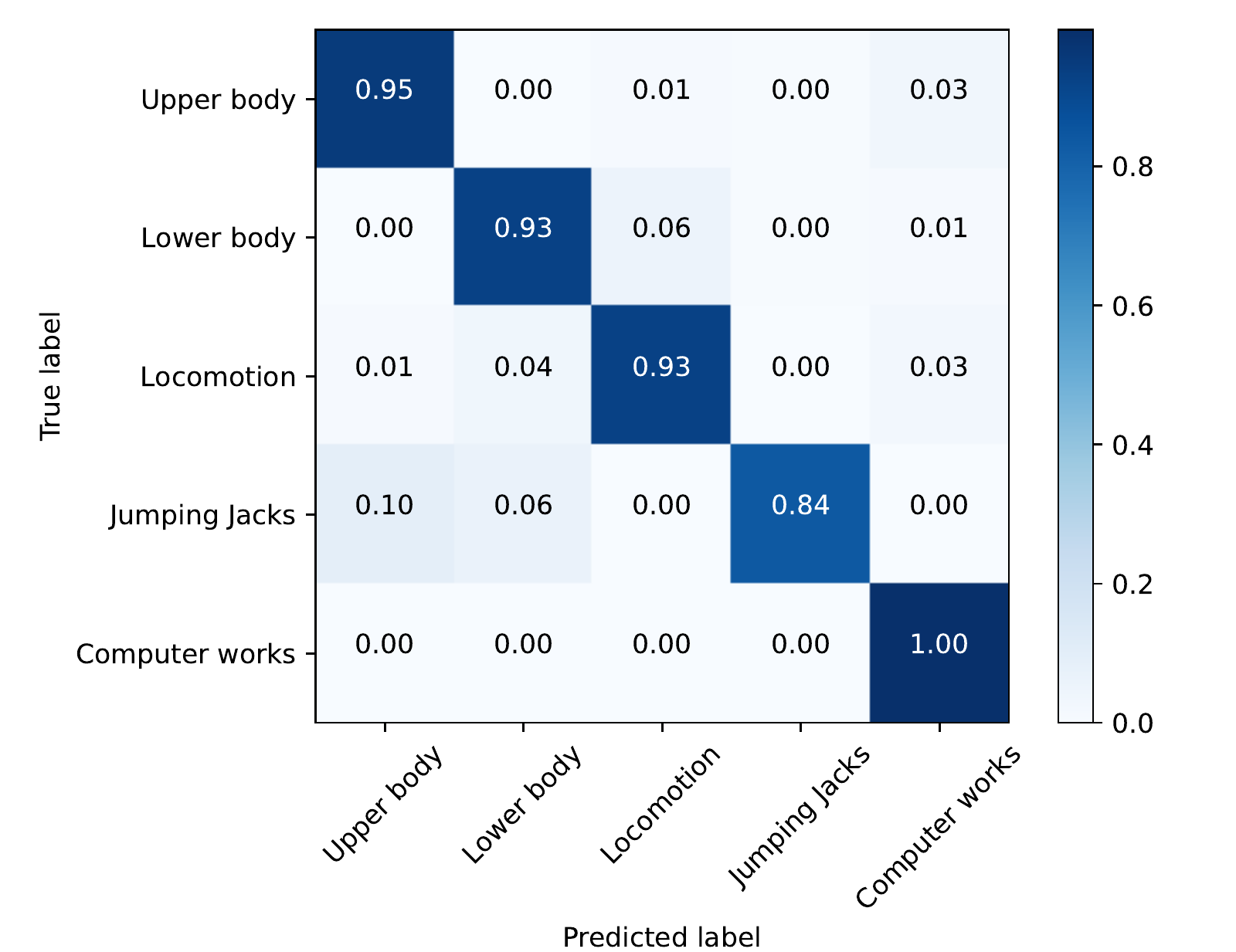}}
	\label{221}
	\subfigure[MARS-v2]{
		\includegraphics[width=6cm]{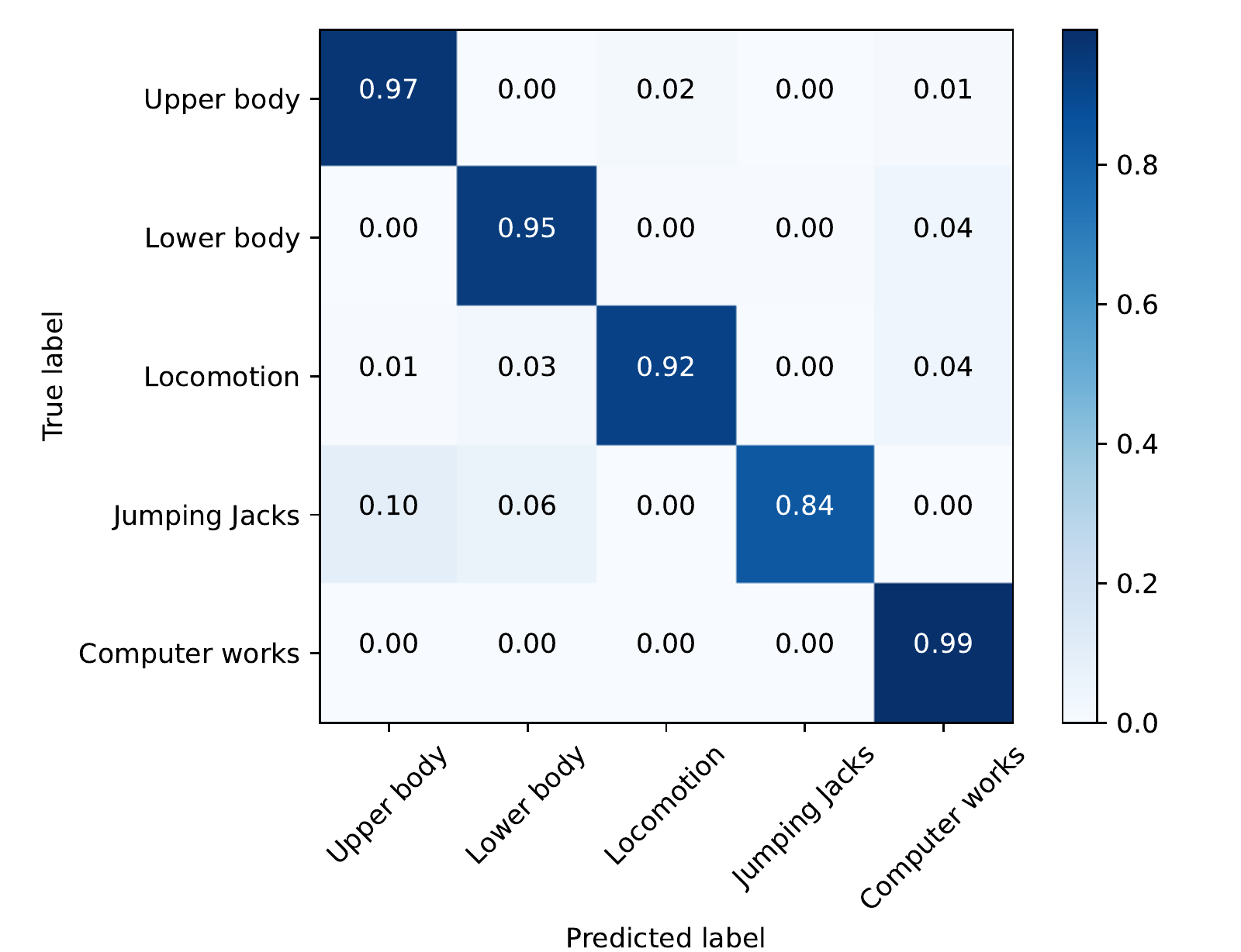}} 
	\label{222}
	\subfigure[MARS-v3]{
		\includegraphics[width=6cm]{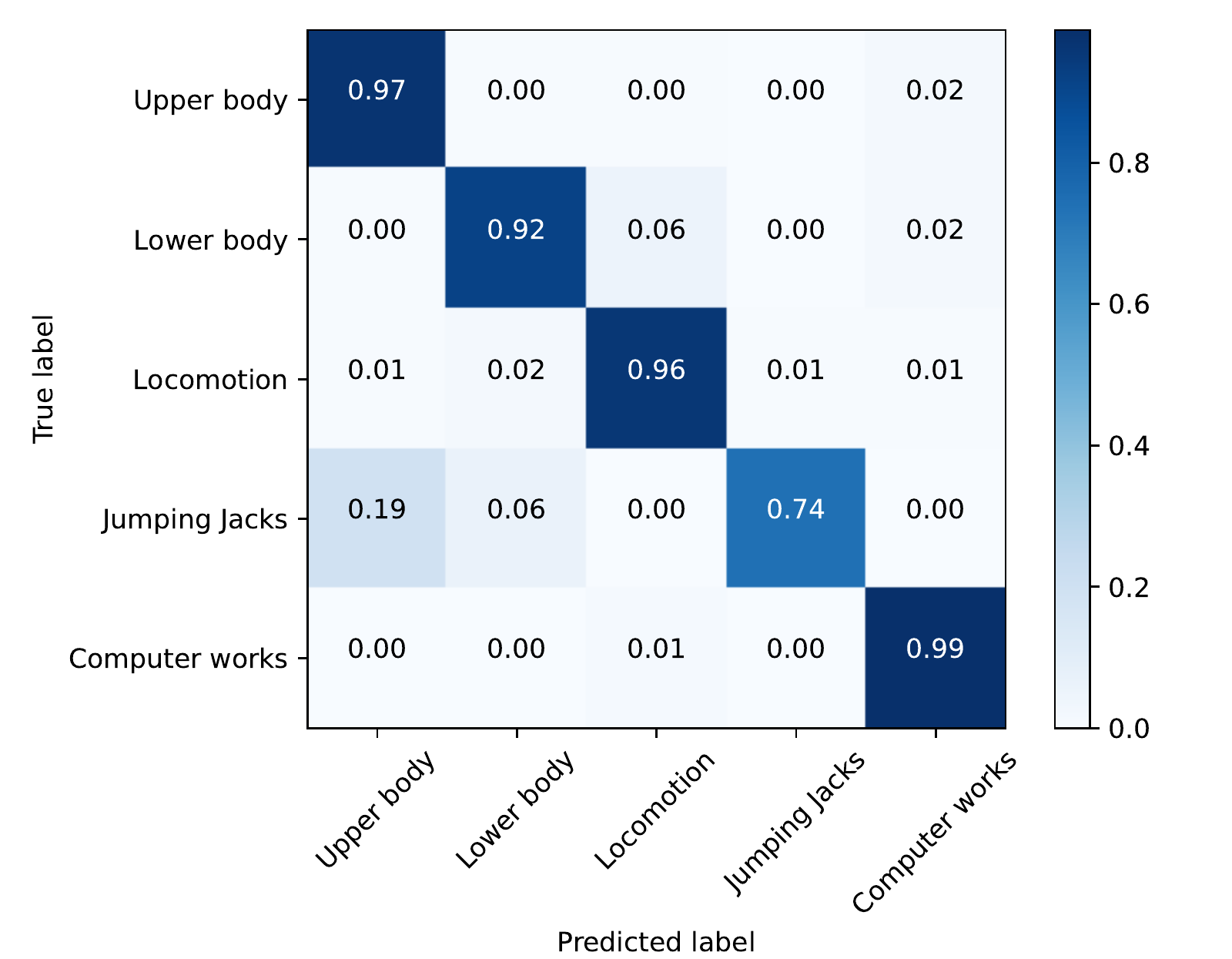}}
	\label{223}

	\caption{Confusion matrices illustration in the second case (6 IMUs). }
	\label{22}
\end{figure*}

\begin{table}[!ht] 
	\caption{\sc Overall Performance Comparison Result}
	\label{demo0}
	\centering
	\begin{tabular}{c|c|c|c|c} 
	    \hline
		\hline
		\textbf{Classifier}  & \textbf{Dataset} & \textbf{Accuracy}   & \textbf{F1-score}   & \textbf{Precision} \\
		\hline
		\textbf{Random Forest} & DIP & \textbf{83.20\%} & \textbf{82.86\%} & \textbf{83.50\%}  \\
		\hline
		\textbf{\multirow{2}{*}{LSTM}} & DIP & \textbf{80.34\%}  & \textbf{82.57\%}  &  \textbf{80.86\%} \\
		\cline{2-5}
		& DIP/AMASS & 73.63\% &  77.28\%  & 74.77\% \\
		\hline 
		\textbf{\multirow{2}{*}{Deep ConvLSTM}} & DIP & 78.33\%  & 79.31\%  &  -\% \\
		\cline{2-5}
		& DIP/AMASS & \textbf{84.80\%} &  \textbf{85.12\%}  & \textbf{-\%} \\
		\hline 
		\textbf{\multirow{2}{*}{Xiao2019}} & DIP & 89.08\%  & 89.16\%  &  -\% \\
		\cline{2-5}
		& DIP/AMASS & \textbf{91.15\%} &  \textbf{91.21\%}  & \textbf{-\%} \\
		\hline 
		\textbf{\multirow{2}{*}{MARS-v1}} & DIP & 91.19\%  & 90.35\%  &  89.87\% \\
		\cline{2-5}
		& DIP/AMASS & \textbf{93.81\%} &  \textbf{92.88\%}  & \textbf{92.00\%} \\
		\hline 
		\textbf{\multirow{2}{*}{MARS-v2}} & DIP & 91.75\%  & 91.77\%  &  92.53\% \\
		\cline{2-5}
		& DIP/AMASS & \textbf{94.50\%} &  \textbf{94.46\%}  & \textbf{94.47\%}  \\
		\hline 
		\textbf{\multirow{2}{*}{MARS-v3}} & DIP & 93.94\%  & 93.93\%  &  94.09\% \\
		\cline{2-5}
		& DIP/AMASS & \textbf{95.81\%} &  \textbf{95.84\%}  & \textbf{95.93\%} \\
		\hline 
		\hline
	\end{tabular}
\end{table}

Tab. \ref{demo0} summarizes the comparison of all algorithms in terms of three performance metrics. It has been shown in Tab. \ref{demo0} that the proposed methods perform favorably over all competing ones when only DIP dataset is available. All three proposed methods can achieve an accuracy of more than 90\% while the comparison algorithm can only achieve a recognition accuracy of 83\%. This result can be explained by the effectiveness of the denoising design and fusion strategies embedded in the proposed methods. Specially, the spatio-temporal feature of the considered dataset can be effectively extracted by the proposed method. On the other hand, the LSTM based algorithm is good at capturing the dependency in real time series data but ignore the spatial information, resulting in decreased HAR results.

What's more, with AMASS dataset, one can perform the knowledge transfer according to Algorithm. \ref{alg1} for all considered deep learning based methods. The results in Tab. \ref{demo0} shows that the knowledge learned from AMASS dataset is beneficial for boosting the performance of deep learning based HAR algorithms except LSTM, demonstrating the adaption of knowledge transfer can bridge the gap between virtual and real IMU datasets.

In order to fully display the test results of various methods, we show the confusion matrix of MARS-v2 and MARS-v3 on the DIP dataset in Fig. \ref{level}. The result shows that the action of Jumping is recognized with the lowest accuracy of 55\%/71\% in MARS-v2 / MARS-v3 without fine-tuning. Jumping is mainly incorrectly labeled as Upper body (39\%/23\%). This error is caused by the complex structure (composite activities) of the Jumping category. For example, the Jumping action in the DIP data set is actually defined as opening and closing jump, which causes the recognition misjudgment between Jumping and Upper body category. Moreover, from the confusion matrix in Fig. \ref{level}, there is an ambiguity between Locomotion and Lower body category. On the other hand, after fine-tuning, the recognition accuracy of Jumping and Locomotion categories has been respectively increased by 10\%/6\% and 14\%/4\%, reaching 65\%/77\% (Jumping) and 96\%/93\% (Locomotion). These result prove that the rich features obtained from the AMASS dataset can boost the performance of the HAR task in real scene.

\subsection{Effect of Sensor Numbers}

For the results shown in Section \ref{opc}, we conducted experiments based on 3 sensors (head, left wrist and right leg). In order to further demonstrate the advantages of using the AMASS dataset, we select more sensors in different locations, which are explained in Tab. \ref{s3456s}. It is noted that when five IMUs are selected, the location of the fifth IMU can be placed on the right wrist or left leg. The specific wearing position and the organization of the collected data are shown in Tab. \ref{s3456s}.

\begin{table}[]
	\caption{\sc Multi-sensor Data Organization Form}
	\setlength{\tabcolsep}{1mm}{
		\label{s3456s}
		\centering
		\begin{tabular}{lllllll}
			\hline
			\hline
			& \textbf{Head} & \textbf{R\_Knee} & \textbf{L\_Wrist} & \textbf{Spine} & \textbf{L\_Knee} & \textbf{R\_Wrist} \\ \hline
			\textbf{3IMU}        & 1$\sim$12     & 13$\sim$24       & 25$\sim$36        &                &                  &                   \\
			\textbf{4IMU}        & 1$\sim$12     & 13$\sim$24       & 25$\sim$36        & 37$\sim$48     &                  &                   \\
			\textbf{5IMU\_knee}  & 1$\sim$12     & 13$\sim$24       & 25$\sim$36        & 37$\sim$48     & 49$\sim$60       &                   \\
			\textbf{5IMU\_wrist} & 1$\sim$12     & 13$\sim$24       & 25$\sim$36        & 37$\sim$48     &                  & 49$\sim$60        \\
			\textbf{6IMU}        & 1$\sim$12     & 13$\sim$24       & 25$\sim$36        & 37$\sim$48     & 49$\sim$60       & 61$\sim$72        \\ \hline
			\hline
		\end{tabular}
	}
\end{table}

In this experiment, the classification performance metrics, including accuracy, F1 scores, and precision, were adopted in the evaluation, and the experimental results were given in Tab. \ref{demo1}.
\begin{table}[!htbp] 
	\caption{\sc Performance Comparison of Proposed Methods with different numbers of IMUs}
	\setlength{\tabcolsep}{1mm}{
		\label{demo1}
		\centering
		\begin{tabular}{c|c|c|c|c|c} 
			\hline
			\hline
			\textbf{Classifier}  & \textbf{IMU Number}  & \textbf{Dataset} & \textbf{Accuaary}   & \textbf{F1-score}   & \textbf{Precision} \\
			
			\hline
			\textbf{\multirow{6}{*}{MARS-v1}} &\multirow{2}{*}{3}  & DIP  & 91.19\%  & 90.35\%  &  89.87\% \\
			\cline{3-6}
			& & \textbf{DIP/AMASS} & \textbf{93.81\%} &  \textbf{92.88\%}  & \textbf{92.00\%} \\
			\cline{2-6}
			&4 & DIP & 91.94\% & 91.02\% & 90.40\% \\
			\cline{2-6}
			&5-knee & DIP & 92.19\% & 91.46\% & 91.13\% \\
			\cline{2-6}
			&5-wrist & DIP & 93.44\% & 92.68\% & 92.47\% \\
			\cline{2-6}
			&\textbf{6} & \textbf{DIP} & \textbf{95.00\%} & \textbf{95.01\%} & \textbf{95.15\%} \\
			\hline			
			
			\textbf{\multirow{6}{*}{MARS-v2}} &\multirow{2}{*}{3}  & DIP  & 91.75\%  & 91.77\%  &  92.53\% \\
			\cline{3-6}
			& & \textbf{DIP/AMASS} & \textbf{94.50\%} &  \textbf{94.46\%}  & \textbf{94.47\%} \\
			\cline{2-6}
			&4 & DIP & 92.69\% & 92.73\% & 92.59\% \\
			\cline{2-6}
			&5-knee & DIP & 94.44\% & 94.47\% & 94.95\% \\
			\cline{2-6}
			&5-wrist & DIP & 94.13\% & 94.19\% & 94.68\% \\
			\cline{2-6}
			&\textbf{6} & \textbf{DIP} & \textbf{95.81\%} & \textbf{95.82\%} & \textbf{95.94\%} \\
			\hline
			
			\textbf{\multirow{6}{*}{MARS-v3}} &\multirow{2}{*}{3}  & DIP  & 93.94\%  & 93.93\%  &  94.09\% \\
			\cline{3-6}
			& & \textbf{DIP/AMASS} & \textbf{95.81\%} &  \textbf{95.84\%}  & \textbf{95.93\%} \\
			\cline{2-6}
			&4 & DIP & 94.00\%& 93.99\%& 94.22\% \\
			\cline{2-6}
			&5-knee & DIP & 95.31\% & 95.26\% & 95.45\% \\
			\cline{2-6}
			&5-wrist & DIP & 95.19\% & 95.21\% & 95.34\% \\
			\cline{2-6}
			& \textbf{6} & \textbf{DIP} & \textbf{96.06\%} & \textbf{96.04\%} & \textbf{96.05\%} \\
			\hline
			\hline			
		\end{tabular}
	}	
\end{table}


In Tab. \ref{demo1}, the notations ``DIP" and ``DIP/AMASS" are referred to the cases that the network training is based solely on the DIP dataset and jointly on the DIP \& AMASS datasets, respectively. Besides, for the case of ``DIP/AMASS", we train the network based on the AMASS and then fine-tune the network parameters with DIP. From Tab. \ref{demo1}, it is clear that the accuracies of HAR improve with an increase in the number of sensors. The MARS-v3 can obtain the best performance for any number of IMUs in both cases, demonstrating the effectiveness of the proposed fusion scheme for the obtained latent features in various domains.

\begin{figure}[!htbp]
	\centering
	\includegraphics[width=8cm]{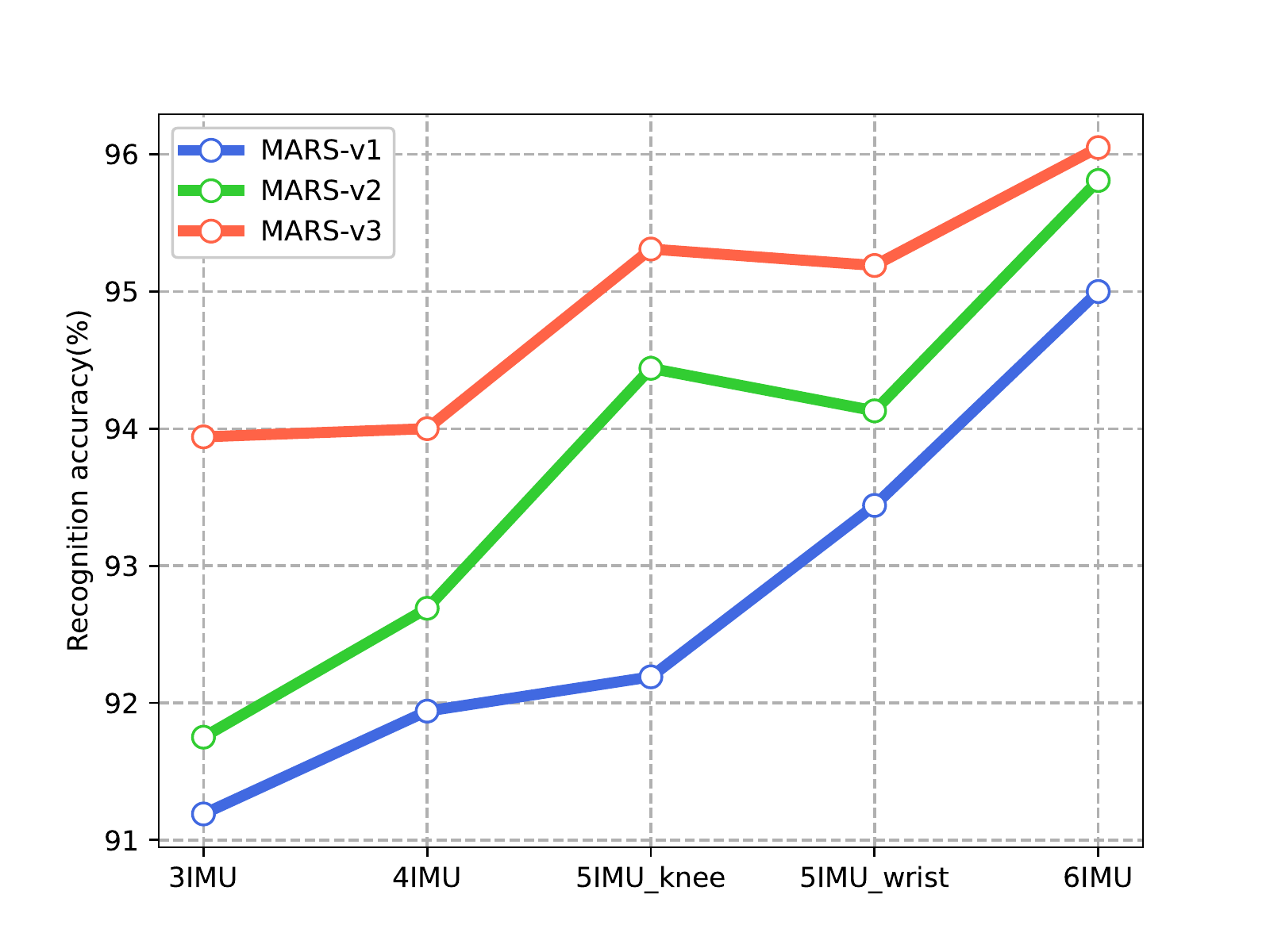}
	
	\caption{HAR accuracy comparison using different numbers of IMUs.}
	\label{Zhexian}
\end{figure}

Besides, from Fig. \ref{Zhexian} and Tab. \ref{demo1}, we can see that the proposed methods' accuracy using six IMUs is higher than 95\%, especially MARS-v3, which can reach 96.06\%. Besides, we can find that the use of 6 IMUs will improve the HAR results in each category compared to that of 3 IMUs. Moreover, as the number of sensors increases, the ambiguity between the Locomotion category and the Lower body category is reduced. However, the ambiguity between the Jumping category and the Upper body category still exists.  

\begin{figure}[!htbp]
	\centering
	\includegraphics[width=8cm]{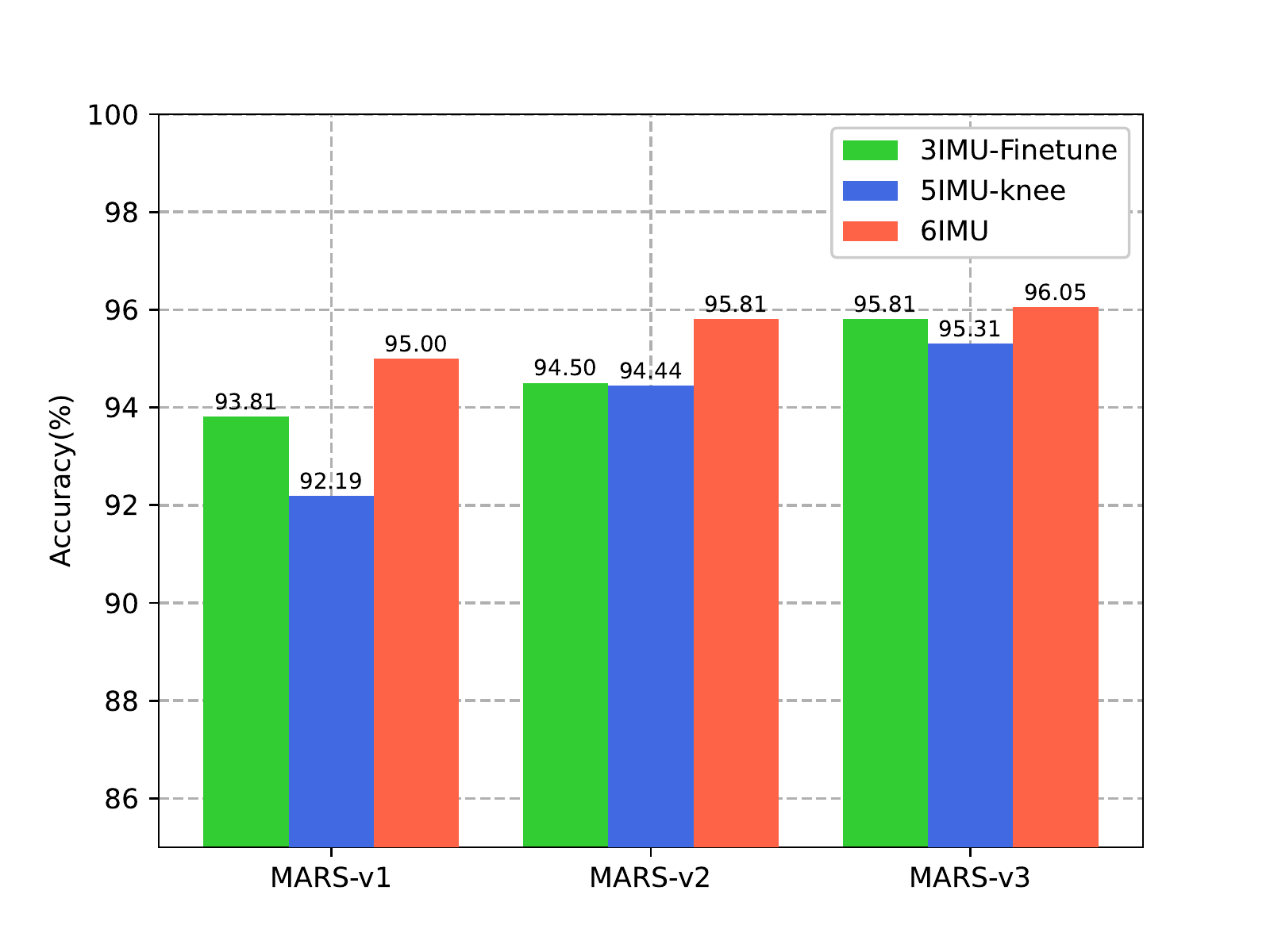}
	
	\caption{Human activity recognition accuracy comparison histogram.}
	\label{histogram}
\end{figure}

Next, we compare the multi-sensor results with fine-tuned results with three IMUs according to Algorithm  \ref{alg1}. Tab. \ref{demo1} and Fig. \ref{histogram} show the comparison between fine-tuned results and the multi-sensor results without the fine-tuning. HAR accuracies of all three proposed methods with three IMUs achieve that with using 6 IMUs. This can be explained by noticing the fact that AMASS contains abundant action features, which encourages effective feature extraction. Besides, combing with the proposed feature fusion and transfer strategies, it would not be surprising that HAR results with only 3 IMUs (with  fine-tuning) can suppress that using 6 IMUs (without fine-tuning).

We will end this section by providing some discussions. It can be seen from the analysis above that since each type of action in the DIP dataset contains many complex composite activities, which results in poor classification performance for some activity, such as Jumping. Increasing the number of wearable sensors seems an effective way to improve the performance of HAR.  For example, when using three sensors, the result of training directly on the DIP is 93.94\%. However, a classifier trained with four sensors has little improvement (0.06\%) in recognition performance. This is because we placed the fourth sensor on the human spine, which contributed relatively little to the HAR in the DIP dataset. After that, when placing the sensor at the knee or wrist, HAR's accuracy will increase to a more considerable extent. The recognition accuracy using five IMUs can reach more than 95\%, which is slightly lower than the result after fine-tuning (95.81\%). As shown in Fig. 8, the recognition accuracy of the proposed methods after fine-tuning reached the result of training using five or six sensors. Therefore, our methods can achieve considerable accuracy while avoiding the intrusion caused by too many sensors, which are favorable to practical applications.

\section{Conclusion and Future Work}
\label{sec:5}
This paper innovatively created a big dataset using virtual IMUs based on the recently released AMASS dataset. To achieve complex HAR in the real scene, we have proposed multiple-domain deep learning methods, consisting of a hybrid feature extractor, comprehensive feature fusion, and transfer module. Extensive experimental results based on the real DIP dataset proved that our proposed methods compare favorably over other competing ones. Significantly, the synthetic AMASS dataset is beneficial for all considered methods to achieve efficient and effective HAR. The proposed methods have dramatically faster training speed and higher recognition accuracy than other comparison methods, reaching 93.81\% and 94.50\% and 95.81\% accuracy. Moreover, combing with the scheme of fine-tuning, the proposed methods with three IMUs can obtain a comparable HAR accuracy obtained from the ones with six IMUs and without fine-tuning, demonstrating again the effectiveness of proposed methods and the novelty of adopting virtual IMU data to realize real HAR.

We will build a more extensive dataset using more virtual IMUs and release it for shared research in our future work. We will also report the HAR results with more datasets and use more methods, such as light-weight neural networks.  

\section*{Acknowledgement}
This work was supported by the National Nature ScienceFoundation of China (NSFC) under Grants: No. 61873163 and No. 61871265.

\bibliography{ref}

\end{document}